\documentclass[acmsmall]{acmart}

\usepackage{tikz}
\usepackage{epstopdf}
\usetikzlibrary{positioning}
\usetikzlibrary{arrows}
\usepackage{subcaption}

\usepackage{comment}

\usepackage{multirow, hhline, bm}
\usepackage{pifont}
\newcommand{\cmark}{\ding{51}}%
\newcommand{\xmark}{\ding{55}}%

\newcommand{\etal}{et~al.~}%

\def\BibTeX{{\rm B\kern-.05em{\sc i\kern-.025em b}\kern-.08emT\kern-.1667em\lower.7ex\hbox{E}\kern-.125emX}}
    
\begin{document}

%
\title{Deep Learning for Iris Recognition: A Survey}

%
\author{Kien Nguyen}
\affiliation{%
  \institution{Queensland University of Technology}
  \country{Australia}
}
\email{nguyentk@qut.edu.au}


\author{Hugo Proen\c{c}a}
\affiliation{%
  \institution{University of Beira Interior, IT: Instituto de Telecomunica\c{c}\~{o}es}
  \country{Portugal}
}
\email{hugomcp@di.ubi.pt}

\author{Fernando Alonso-Fernandez}
\affiliation{%
  \institution{Halmstad University}
  \country{Sweden}
}
\email{feralo@hh.se}

%
\renewcommand{\shortauthors}{K. Nguyen, H. Proen\c{c}a, F. Alonso-Fernandez}

%
\begin{abstract}
ABSTRACT


\noindent In this survey, we provide a comprehensive review of more than 200 papers, technical reports, and GitHub repositories published over the last 10 years on the recent developments of deep learning techniques for iris recognition, covering broad topics on algorithm designs, open-source tools, open challenges, and emerging research. First, we conduct a comprehensive analysis of deep learning techniques developed for two main sub-tasks in iris biometrics: segmentation and recognition. Second, we focus on deep learning techniques for the robustness of iris recognition systems against presentation attacks and via human-machine pairing. Third, we delve deep into deep learning techniques for forensic application, especially in post-mortem iris recognition. Fourth, we review open-source resources and tools in deep learning techniques for iris recognition. Finally, we highlight the technical challenges, emerging research trends, and outlook for the future of deep learning in iris recognition.

\end{abstract}

%
%


%

%

%
\maketitle

\section{Introduction}

The human iris is a sight organ that controls the amount of light reaching the retina, by changing the size of the pupil. The texture of the iris is fully developed before birth, its minutiae do not depend on genotype, it stays relatively stable across lifetime (except for disease- and normal aging-related biological changes), and it may even be used for forensic identification  shortly after subject's death~\cite{IrisStructure,DaugmanInformationTheory,Trokielewicz_TIFS_2019}. 

In terms of its information theory-related properties, the iris texture has an extremely  high randotypic randomness, and is stable (permanent) over time, providing an exceptionally high entropy per mm.$^2$ that justifies its higher discriminating power, when compared to other biometric modalities (e.g., face or fingerprint). 
The iris' collectability is another feature of interest and has been the subject of discussion over the last years: while it can be acquired using  commercial off-the-shelf (COTS) hardware, either handheld or stationary, data can be even collected from at-a-distance, up to tens of meters away from the subjects \cite{IAADsurvey}. Even though commercial visible-light (RGB) cameras are able to image the iris, the near infrared-based (NIR) sensing dominates in most applications, due to a better visibility of iris texture for darker eyes, rich in melanin pigment, which is characterized by lower light absorption in NIR spectrum compared to shorter wavelengths. In addition, NIR wavelengths are barely perceivable by the human eye, which augment users' comfort, and avoids pupil contraction/dilation that would appear under visible light.

 


A seminal work by John Daugman brought to the community the Gabor filtering-based approach that became the dominant approach for iris recognition  \cite{244676,IrisCode,DaugmanCollisionAvoidance}. Even though subsequent solutions to iris image encoding and matching appeared, 
the IrisCodes approach is still dominant  due to its ability to effectively search in massive databases with a minimal probability of false matches, at extreme time performance. By considering binary words, pairs of signatures are matched using XOR parallel-bit logic at lightening speed, enabling millions of comparisons/second per processing core. Also, most of the methods that outperformed the original techniques in terms of effectiveness do not work under the \emph{one-shot learning} paradigm, assume multiple observations of each class to obtain appropriate decision boundaries, and - most importantly - have encoding/matching steps with time complexity that forbid their use in large environments (in particular, for \emph{all-against-all} settings).

In short, Daugman's algorithm encodes the iris image into a binary sequence of 2,048 bits by filtering the iris image with a family of Gabor kernels. The varying pupil size is rectified by the Cartesian-to-polar coordinate system transformation, to end up with an image representation of canonical size, guarantying identical structure of the iris code independently of the iris and pupil size. This makes possible to use the Hamming Distance (HD) to measure the similarity between two iris codes \cite{DaugmanCollisionAvoidance}.
Its low false match rate at acceptable false non-match rates is the key factor behind the success of global-scale iris recognition installments, such as the national person identification and border security program Aadhaar program in India (with over 1.2 billion pairs of irises enrolled) \cite{Aadhaar}, the Homeland Advanced Recognition Technology (HART) in the US (up to 500 million identities) \cite{USHART}, or the NEXUS system, designed to speed up border crossings for low-risk and pre-approved travelers moving between Canada and the US.

Deep learning-based methods, in particular using various Convolutional Neural Network architectures, have been driving remarkable improvements in many computer vision applications over the last decade. In terms of biometrics technologies, it's not surprising that iris recognition has also seen an increasing adoption of purely data-driven approaches at all stages of the recognition pipeline: from preprocessing (such as off-axis gaze correction), segmentation, encoding to matching. 
Interestingly, however, the impact of deep learning on the various stages of iris recognition pipeline is uneven. One of the primary goals of this survey paper is to assess where deep learning helped in achieving highly performance and more secure systems, and which procedures did not benefit from more complex modeling. 

The remainder of the paper is structured as follows. Section~\ref{sec:segmentation} and ~\ref{sec:recognition} review the application of deep learning in two main stages of the recognition pipeline: segmentation and recognition (encoding and comparison). Section~\ref{sec:presentationattack} and ~\ref{sec:forensic} analyze the state of the art of deep learning-based approaches in two applications: Presentation Attack Detection (PAD) and Forensic. Section~\ref{sec:humanpairing} investigates how human and machine can pair to improve deep learning based iris recognition. Section~\ref{sec:periocular} focuses on approaches in less controlled environments of iris and periocular analysis. Section~\ref{sec:opensources} reviews public resources and tools available in the deep learning based iris recognition domain. Section~\ref{sec:open} focuses on the future of deep learning for iris recognition with discussion on emerging research directions in different aspects of iris analysis. The paper in concluded in Section~\ref{sec:Conclusion}.

\section{Deep Learning-Based Iris Segmentation}
\label{sec:segmentation}



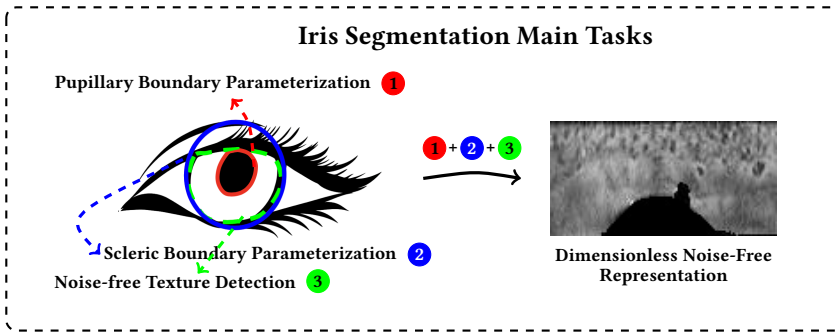
\begin{figure}[ht!]
\begin{center}
  \begin{tikzpicture}

    \draw (0,0) node(n1)  {\includegraphics[width=3.5 cm]{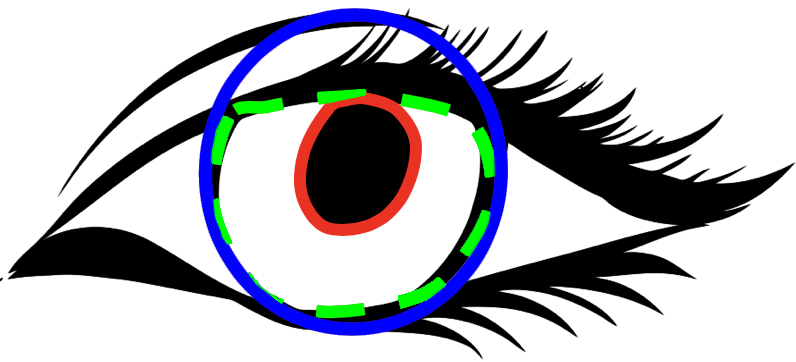}};                       
    
    \def\deltaX{0.25}

   \draw [dashed, very thick,blue, ->] (-0.9,0.25) .. controls (-2.5, -0.5) .. (-2.0,-1);	    
\draw (0.0,-1) node {\scriptsize{\textbf{Scleric Boundary Parameterization}}};        
\draw [fill, blue](2.25,-1.0)circle(0.15);
\draw (2.25, -1.0) node [white] {\scriptsize{\textbf{2}}};          

   \draw [dashed, very thick,green, ->] (-0.1,-0.5) .. controls (-0.3, -0.75) .. (-0.7,-1.25);	    
\draw (-1.0,-1.35) node {\scriptsize{\textbf{Noise-free Texture Detection}}};       
\draw [fill, green](0.9,-1.35)circle(0.15);
\draw (0.9, -1.35) node [black] {\scriptsize{\textbf{3}}};    

   \draw [dashed, very thick,red, ->] (0.05,0.15) .. controls (0.0, 0.75) .. (-0.25,1.05);	    
\draw (-0.5,1.25) node {\scriptsize{\textbf{Pupillary Boundary Parameterization}}};          
\draw [fill, red](1.9,1.25)circle(0.15);
\draw (1.9, 1.25) node [black] {\scriptsize{\textbf{1}}};    
             
         \draw (5.5,0) node(n1)  {\includegraphics[width=3.0 cm]{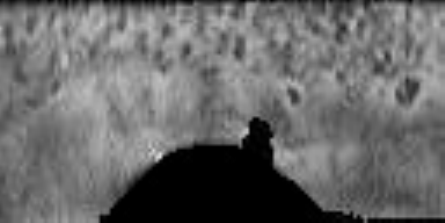}};

     \draw [thick, dashed, rounded corners ] (-3.25, 2.25) rectangle (1.85+6.1,-2.0);       

   \draw (3,1.85) node {\textbf{Iris Segmentation Main Tasks}};              

   \draw (5.5,-1.0) node {\scriptsize{\textbf{Dimensionless Noise-Free }}};              
   \draw (5.5,-1.3) node {\scriptsize{\textbf{Representation}}};    
           
         \def\shiftY{-3.75}
     \def\shiftX{1.0}     
    \draw [->, very thick] (1.3+\shiftX,3.75+\shiftY) .. controls (2.0+\shiftX, 3.85+\shiftY) .. (2.6+\shiftX, 3.75+\shiftY);

\draw [fill, blue](2.95,0.4)circle(0.15);
\draw (2.95, 0.4) node [white] {\scriptsize{\textbf{2}}};     

\draw [fill, red](2.45,0.4)circle(0.15);
\draw (2.45, 0.4) node [black] {\scriptsize{\textbf{1}}};     

\draw (2.7, 0.4) node [black] {\scriptsize{\textbf{+}}};     
\draw (3.2, 0.4) node [black] {\scriptsize{\textbf{+}}};     

\draw [fill, green](3.45,0.4)circle(0.15);
\draw (3.45, 0.4) node [black] {\scriptsize{\textbf{3}}};         
    
\end{tikzpicture}
    \caption{Three main tasks typically associated to \emph{iris segmentation}: 1) parameterization of the pupillary (inner) boundary; 2) parameterization of the scleric (outer) boundary; and 3) discrimination between the unoccluded (noise-free) and occluded (noisy) regions inside the iris ring. Such pieces of information are further used to obtain dimensionless polar representations of the iris texture, where feature extraction methods typically operate. }
    \label{fig:Segmentation}
    \end{center}
\end{figure}

The segmentation of the iris is seen as an extremely challenging problem. As illustrated in Fig.~\ref{fig:Segmentation}, segmenting the iris involves essentially three tasks: detect and parameterize the inner (pupillary) and outer (scleric) biological boundaries of the iris and also to locally discriminate between the noise-free/noisy regions inside the iris ring, which should be subsequently used in the feature encoding and matching processes. 

This  problem has motivated numerous research works for decades. From the pioneering integro-differential operator~\cite{244676} up to subsequent handcrafted techniques based in active contours and neural networks (e.g.,~\cite{4586378},~\cite{5156505},~\cite{5272431} and~\cite{4510759}) a long road has been traveled in this problem. Regardless an obvious evolution in the effectiveness of such techniques, they all face particular difficulties in case of heavily degraded data. Images are frequently motion-blurred, poor focused, partially occluded and off-angle. Additionally, in case of visible light data, severe reflections from the environments surrounding the subjects are visible, and even augment the difficulties of the segmentation task.      

Recently, as in many other computer vision tasks, DL-based frameworks have been advocated as providing consistent advances over the state-of-the-art for the iris segmentation problem, with numerous models being proposed. A cohesive perspective of the most relevant recent DL-based methods is given in Table~\ref{tab:Segmentation}, with the techniques appearing in chronographic (and then alphabetical) order. The type of data each model aims to handle is given in the ''\emph{Data}'' column, along with the datasets where the corresponding experiments were carried out and a summary of the main characteristics of each proposal (''\emph{Features}'' column). Here, considering that models were empirically validated in completely heterogeneous ways and using very different metrics, we decided not to include the summary performance of each model/solution.      

Schlett \emph{et} al.~\cite{8411221} provided a multi-spectral analysis to improve iris segmentation accuracy in visible wavelengths by preprocessing data before the actual segmentation phase, extracting multiple spectral components in form of RGB color channels. Even though this approach does propose a DL-based framework, the different versions of the input could be easily used to feed  DL-based models, and augment the robustness to non-ideal data. Chen \emph{et} al.~\cite{8715779}  used CNNs that include dense blocks, referred to as a dense-fully convolutional network (DFCN), where the encoder part  consists of dense blocks, and the decoder counterpart obtains the segmentation masks via transpose convolutions. Hofbauer \emph{et} al.~\cite{CNNHT} parameterize the iris boundaries based on segmentation maps yielding from a CNN, using a a cascaded architecture with four RefineNet units, each directly connecting to one Residual net. Huynh \emph{et} al.~\cite{9022251} discriminate between three distinct eye regions with a DL model, and removes incorrect areas with heuristic filters. The proposed architecture is based on the encoder-decoder model, with depth-wise convolutions used to reduce the computational cost. Roughly at the same time, Li \emph{et} al.~\cite{s21041434} described the \emph{Interleaved Residual U-Net} model for semantic segmentation and iris mask synthesis. In this work, unsupervised techniques (K-means clustering) were used to create intermediary pictorial representations of the ocular region, from where saliency points deemed to belong to the iris boundaries were found. Kerrigan \emph{et} al.~\cite{8987299} assessed the performance of four different convolutional architectures designed for semantic segmentation. Two of these models were based in dilated convolutions, as proposed by Yu and Koltun~\cite{yu2016multiscale}.  Wu and Zhao~\cite{8822410} described the Dense U-Net model,  that combines dense layers to the U-Net network. The idea is to take advantage of the reduced set of parameters of the dense U-Net,  while keeping the semantic segmentation capabilities of U-Net. The proposed model integrates dense connectivity into U-Net contraction and expansion paths. Compared with traditional CNNs, this model is claimed to reduce learning redundancy and enhance information flow, while keeping controlled the number of parameters of the model. Wei \emph{et} al.~\cite{8744291} suggested to perform \emph{dilated convolutions}, which is claimed to obtain more consistent global features. In this setting, convolutional kernels are not continuous, with zero-values being artificially inserted between each non-zero position, increasing the receptive field without  augmenting the number of parameters of the model.

\begin{table*}[h]
    \centering
     \caption{Cohesive comparison of the most relevant DL-based iris segmentation methods (NIR: \emph{near-infrared}; VW: \emph{visible wavelength}). Methods are listed in chronological (and then alphabetical) order.}
    \begin{tabular}{|p{1.88cm}|p{.45cm}|p{.45cm}|p{.45cm}|p{4.2cm}|p{4.4cm}|}\hline
         \multirow{2}{*}{ \textbf{\scriptsize{Method}}}  &  \multirow{2}{*}{ \textbf{\scriptsize{Year}}}  &\multicolumn{2}{c|}{ \textbf{\scriptsize{Data}}}&   \multirow{2}{*}{\textbf{\scriptsize{Datasets}}} &   \multirow{2}{*}{\textbf{\scriptsize{Features}}}  \\ \cline{3-4}
         
         & & \textbf{\scriptsize{NIR}} & \textbf{\scriptsize{VW}} &  & \\ \hline

        \scriptsize{Schlett \emph{et} al.~\cite{8411221}} & \scriptsize{2018} & \xmark & \cmark & \scriptsize{MobBIO}  & \scriptsize{Preprocessing (combines different possibilities of the input RGB channels)}  \\ \hline
           
        \scriptsize{Trokielewicz and Czajka~\cite{Trokielewicz_IWBF_2018}} & \scriptsize{2018} & \cmark & \cmark & \scriptsize{Warsaw-Post-Mortem v1.0}  & \scriptsize{Fine-tuned CNN (SegNet)} \\ \hline
        
	\scriptsize{Chen \emph{et} al.~\cite{8715779}} & \scriptsize{2019} & \cmark & \cmark & \scriptsize{CASIA-Irisv4-Interval, IITD, UBIRIS.v2}  & \scriptsize{Dense CNN}  \\ \hline

	\scriptsize{Hofbauer \emph{et} al.~\cite{CNNHT} } & \scriptsize{2019} & \cmark & \xmark & \scriptsize{IITD, CASIA-Irisv4-Interval, ND-Iris-0405}  & \scriptsize{Cascaded architecture of four RefineNet, each connecting to one Residual net}  \\ \hline

	\scriptsize{Huynh \emph{et} al.~\cite{9022251}} & \scriptsize{2019} & \cmark & \xmark & \scriptsize{OpenEDS}  & \scriptsize{MobileNetV2 
	+ heuristic filtering postproc. 
	}  \\ \hline	        

        \scriptsize{Li \emph{et} al.~\cite{Li2019} } & \scriptsize{2019} & \cmark & \xmark & \scriptsize{CASIA-Iris-Thousand}  & \scriptsize{Faster-R-CNN (ROI detection)}  \\ \hline	
	
	\scriptsize{Kerrigan \emph{et} al.~\cite{8987299}} & \scriptsize{2019} & \cmark & \cmark & \scriptsize{CASIA-Irisv4-Interval, BioSec, ND-Iris-0405, UBIRIS.v2, Warsaw-Post-Mortem v2.0, ND-TWINS-2009-2010}  & \scriptsize{Resent + Segnet (with dilated convolutions)}  \\ \hline
                                                        
	 \scriptsize{Wu and Zhao~\cite{8822410}} & \scriptsize{2019} & \cmark & \cmark & \scriptsize{CASIA-Irisv4-Interval, UBIRIS.v2}  & \scriptsize{Dense-U-Net (dense layers + U-Net) 
	 }  \\ \hline
                                                                	
        \scriptsize{Wei \emph{et} al.~\cite{8744291}} & \scriptsize{2019} & \cmark & \cmark & \scriptsize{CASIA-Iris4-Interval, ND-IRIS-0405, UBIRIS.v2}  & \scriptsize{U-Net with dilated convolutions}  \\ \hline

    \scriptsize{Fang and Czajka~\cite{Fang_IJCB_2020}} & \scriptsize{2020} & \cmark & \cmark & \scriptsize{ND-Iris-0405, CASIA, BATH, BioSec, UBIRIS, Warsaw-Post-Mortem v1.0 \& v2.0}  & \scriptsize{Fine-tuned CC-Net~\cite{CC-Net}}  \\ \hline

	\scriptsize{Ganeva and Myasnikov~\cite{9271541}} & \scriptsize{2020} & \cmark & \xmark & \scriptsize{MMU}  & \scriptsize{U-Net, LinkNet, and FC-DenseNet (performance comparison)}  \\ \hline
	
             \scriptsize{Jalilian \emph{et} al.~\cite{9210983} } & \scriptsize{2020} & \cmark & \xmark & \scriptsize{}  & \scriptsize{
             RefineNet + morphological postprocessing}  \\ \hline
       
 	\scriptsize{Sardar \emph{et} al.~\cite{9274419}} & \scriptsize{2020} & \cmark & \cmark & \scriptsize{CASIA-Irisv4-Interval, IITD, UBIRIS.v2}  & \scriptsize{Squeeze-Expand module + active learning (interactive segmentation)}  \\ \hline

	\scriptsize{Trokielewicz \emph{et} al.~\cite{Trokielewicz_IMAVIS_2020} } & \scriptsize{2020} & \cmark & \cmark & \scriptsize{ND-Iris-0405, CASIA, BATH, BioSec, UBIRIS, Warsaw-Post-Mortem v1.0 \& v2.0}  & \scriptsize{Fined-tuned SegNet~\cite{Badrinarayanan_TPAMI_2017}}  \\ \hline                                                         

        \scriptsize{Wang et al~\cite{9054353}} & \scriptsize{2020} & \cmark & \cmark & \scriptsize{CASIA-Iris-M1-S1/S2/S3, MICHE-I}  & \scriptsize{Hourglass network}  \\ \hline

	 \scriptsize{Wang \emph{et} al.~\cite{IrisParseNet} } & \scriptsize{2020} & \cmark & \cmark & \scriptsize{CASIA-v4-Distance, UBIRIS.v2, MICHE-I}  & \scriptsize{U-Net + multi-task attention net + postproc. (probabilistic masks priors \& thresholding)
	 }  \\ \hline
                                                
	\scriptsize{Li \emph{et} al.~\cite{s21041434}} & \scriptsize{2021} & \cmark & \xmark & \scriptsize{CASIA-Iris-Thousand}  & \scriptsize{IRU-Net network}  \\ \hline     
	
        \scriptsize{Wang \emph{et} al.~\cite{8818661}} & \scriptsize{2021} & \xmark & \cmark & \scriptsize{Online Video Streams and Internet Videos}  & \scriptsize{U-Net and Squeezenet to iris segmentation and detect eye closure }  \\ \hline
        
	\scriptsize{Kuehlkamp \etal \cite{Kuehlkamp_WACV_2022} } & \scriptsize{2022} & \cmark & \cmark & \scriptsize{
	ND-Iris-0405, CASIA, BATH, BioSec, UBIRIS, Warsaw-Post-Mortem v2.0}  & \scriptsize{Fined-tuning of Mask-RCNN architecture}  \\ \hline  
	
    \end{tabular}
    \label{tab:Segmentation}
\end{table*}

More recently, Ganeva and Myasnikov~\cite{9271541} compared the effectiveness of three convolutional neural network architectures (U-Net, LinkNet, and FC- DenseNet), determining the optimal parameterization for each one.  Jalilian \emph{et} al.~\cite{9210983} introduced a scheme to compensate for texture deformations caused by the off-angle distortions, re-projecting the off-angle images back to frontal view. The used architecture is a variant of RefineNet~\cite{lin2016refinenet}, which provides high-resolution prediction, while preserving the boundary information (required for parameterization purposes).

The idea of interactive learning for iris segmentation was suggested by Sardar \emph{et} al.~\cite{9274419}, describing an interactive variant of U-Net that includes Squeeze Expand modules. Trokielewicz \emph{et} al.~\cite{Trokielewicz_IMAVIS_2020} used DL-based iris segmentation models to extract highly irregular iris texture areas in post-mortem iris images. They used a pre-trained SegNet model, fine-tuned with a database of cadaver iris images. Wang \emph{et} al.~\cite{9054353} (further extended in~\cite{wang2019joint}) described a lightweight deep convolutional neural network specifically designed for iris segmentation of degraded images acquired by handheld devices.  The proposed approach jointly obtains the segmentation mask and parameterized pupillary/limbic boundaries of the iris. 

Observing that edge-based information is extremely sensitive to be obtained in degraded data,  Li \emph{et} al.~\cite{Li2019} presented an hybrid method that combines edge-based information to deep learning frameworks. A compacted Faster R-CNN-like architecture was used to roughly detect the eye and define the initial region of interest, from where the pupil is further located using a Gaussian mixture model.  Wang \emph{et} al.~\cite{8818661} trained a deep convolutional neural network(DCNN) that automatically extracts the iris and pupil pixels of each eye from input images. This work combines the power of U-Net and SqueezeNet to obtain a compact CNN suitable for real time mobile applications. Finally, Wang \emph{et} al.~\cite{IrisParseNet} parameterize both the iris mask and the inner/outer iris boundaries jointly, by actively modeling such information into a unified multi-task network. 

A final word is given to \emph{segmentation-less} techniques. Assuming that the accurate segmentation of the iris boundaries is one of the hardest phases of the whole recognition chain and the main source for recognition errors, some recent works have been proposing to perform biometrics recognition in non-segmented or roughly segmented data~\cite{IRINA}\cite{Segmentation-Less}. Here, the idea is to use the remarkable discriminating power of DL-frameworks to perceive the agreeing patterns between pairs of images, even on such \emph{segmentation-less} representations.

\section{Deep Learning-Based Iris Recognition} 
\label{sec:recognition}


\subsection{Deep Learning Models as a Feature Extractor} 

\begin{figure}[ht!]
\begin{center}
  \begin{tikzpicture}
  
   \draw (2.5,1.5) node {\textbf{DL-based Feature Representation}};              
  
       \draw [thick, dashed, rounded corners ] (-3.25, 2.0) rectangle (1.85+6.1,-2.0);       
  
    \draw (-0.25,0) node(n1)  {\includegraphics[width=3.5 cm]{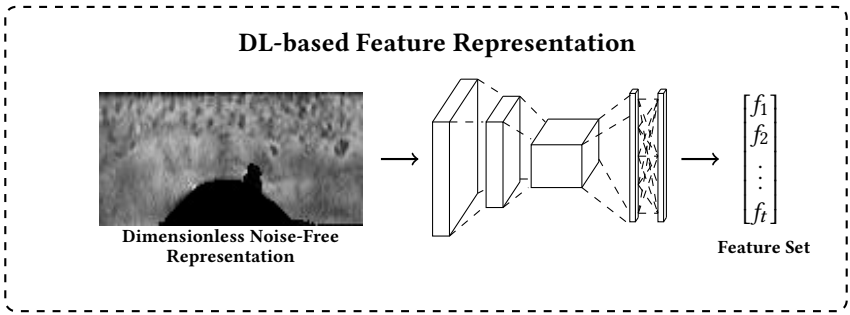}};                       
   
   \draw (-0.25,-1.0) node {\scriptsize{\textbf{Dimensionless Noise-Free }}};              
   \draw (-0.25,-1.3) node {\scriptsize{\textbf{Representation}}};

   \draw (6.85,-1.15) node {\scriptsize{\textbf{Feature Set}}};              

  \draw [->, thick] (1.75, 0) -- (2.25, 0);      
  \draw [->, thick] (5.75, 0) -- (6.25, 0);        

      \begin{scope}[scale=0.75]

\def\globalX{2.5}
\def\globalY{-1.35}

  \def\sizeY{2}  \def\posX{0.75+\globalX}   \def\posY{2+\globalY}   \def\depth{0.3}\def\perspectiveX{0.5}  \def\perspectiveY{\perspectiveX*1.5}  
  \draw (\posX,\posY) rectangle (\posX+\depth,\posY-\sizeY);     
  \draw (\posX+\perspectiveX, \posY+\perspectiveY) -- (\posX+\perspectiveX+\depth,  \posY+\perspectiveY);
 \draw (\posX+\perspectiveX+\depth,  \posY+\perspectiveY) -- (\posX+\perspectiveX+\depth,  \posY-\sizeY+\perspectiveY);  
    \draw (\posX+\perspectiveX,  \posY+\perspectiveY) -- (\posX,  \posY);    
   \draw (\posX+\perspectiveX+\depth,  \posY-\sizeY+\perspectiveY) -- (\posX+\depth,  \posY-\sizeY);      
   \draw (\posX+\perspectiveX+\depth,  \posY+\perspectiveY) -- (\posX+\depth,  \posY);         
   
   \def\posXNextLayer{0.5}
   \draw [dashed] (\posX+\depth,\posY-\sizeY) -- (\posX+0.9, \posY-1.5);      
   \draw [dashed] (\posX+\depth+\perspectiveX,\posY-\sizeY+0.75) -- (\posX+0.9, \posY-1.2);         
      \draw [dashed] (\posX+\depth,\posY) -- (\posX+0.9, \posY+0.0);      
      \draw [dashed] (\posX+\depth+\perspectiveX,\posY+\perspectiveY) -- (\posX+0.7+\perspectiveX, \posY+0.5);            

  \def\sizeY{1.5}  \def\posX{1.7+\globalX}   \def\posY{2+\globalY}   \def\depth{0.3}\def\perspectiveX{0.3}  \def\perspectiveY{\perspectiveX*1.5}  
  \draw (\posX,\posY) rectangle (\posX+\depth,\posY-\sizeY);     
  \draw (\posX+\perspectiveX, \posY+\perspectiveY) -- (\posX+\perspectiveX+\depth,  \posY+\perspectiveY);
 \draw (\posX+\perspectiveX+\depth,  \posY+\perspectiveY) -- (\posX+\perspectiveX+\depth,  \posY-\sizeY+\perspectiveY);  
    \draw (\posX+\perspectiveX,  \posY+\perspectiveY) -- (\posX,  \posY);    
   \draw (\posX+\perspectiveX+\depth,  \posY-\sizeY+\perspectiveY) -- (\posX+\depth,  \posY-\sizeY);      
   \draw (\posX+\perspectiveX+\depth,  \posY+\perspectiveY) -- (\posX+\depth,  \posY);

   \def\posXNextLayer{0.5}
   \draw [dashed] (\posX+\depth,\posY-\sizeY) -- (\posX+0.75, \posY-1.15);      
   \draw [dashed] (\posX+\depth+\perspectiveX,\posY-\sizeY+0.5) -- (\posX+0.8, \posY-1.0);         
    \draw [dashed] (\posX+\depth,\posY) -- (\posX+0.8, \posY-0.3);      
      \draw [dashed] (\posX+\depth+\perspectiveX,\posY+\perspectiveY) -- (\posX+0.75+\perspectiveX, \posY);     

  \def\sizeY{0.75}  \def\posX{2.5+\globalX}   \def\posY{1.6+\globalY}   \def\depth{0.9}\def\perspectiveX{0.3}  \def\perspectiveY{\perspectiveX*1.5}  
  \draw (\posX,\posY) rectangle (\posX+\depth,\posY-\sizeY);     
  \draw (\posX+\perspectiveX, \posY+\perspectiveY) -- (\posX+\perspectiveX+\depth,  \posY+\perspectiveY);
 \draw (\posX+\perspectiveX+\depth,  \posY+\perspectiveY) -- (\posX+\perspectiveX+\depth,  \posY-\sizeY+\perspectiveY);  
    \draw (\posX+\perspectiveX,  \posY+\perspectiveY) -- (\posX,  \posY);    
   \draw (\posX+\perspectiveX+\depth,  \posY-\sizeY+\perspectiveY) -- (\posX+\depth,  \posY-\sizeY);      
   \draw (\posX+\perspectiveX+\depth,  \posY+\perspectiveY) -- (\posX+\depth,  \posY);         
   
   \def\posXNextLayer{0.5}
   \draw [dashed] (\posX+\depth,\posY-\sizeY) -- (\posX+1.75, \posY-1.4);      
   \draw [dashed] (\posX+\depth+\perspectiveX,\posY-\sizeY+0.5) -- (\posX+1.75, \posY-1.2);         
      \draw [dashed] (\posX+\depth,\posY) -- (\posX+1.75, \posY+0.5);      
      \draw [dashed] (\posX+\depth+\perspectiveX,\posY+\perspectiveY) -- (\posX+1.5+\perspectiveX, \posY+0.9);

  \def\sizeY{2.25}  \def\posX{4.25+\globalX}   \def\posY{+2.5+\globalY}   \def\depth{0.1}\def\perspectiveX{0.05}  \def\perspectiveY{\perspectiveX*1.5}  
  \draw (\posX,\posY) rectangle (\posX+\depth,\posY-\sizeY);     
  \draw (\posX+\perspectiveX, \posY+\perspectiveY) -- (\posX+\perspectiveX+\depth,  \posY+\perspectiveY);
 \draw (\posX+\perspectiveX+\depth,  \posY+\perspectiveY) -- (\posX+\perspectiveX+\depth,  \posY-\sizeY+\perspectiveY);  
    \draw (\posX+\perspectiveX,  \posY+\perspectiveY) -- (\posX,  \posY);    
   \draw (\posX+\perspectiveX+\depth,  \posY-\sizeY+\perspectiveY) -- (\posX+\depth,  \posY-\sizeY);      
   \draw (\posX+\perspectiveX+\depth,  \posY+\perspectiveY) -- (\posX+\depth,  \posY);

   \def\posXNextLayer{0.5}
   \def\posY{+1.9+\globalY}
   
    \draw [dashed] (\posX+0.15, 0.5+\posY) -- (\posX+\posXNextLayer, -0.5+\posY);   
    \draw [dashed] (\posX+0.15, 0+\posY) -- (\posX+\posXNextLayer, -0.5+\posY);   
    \draw [dashed] (\posX+0.15, -0.5+\posY) -- (\posX+\posXNextLayer, -0.5+\posY);
    \draw [dashed] (\posX+0.15, -1.0+\posY) -- (\posX+\posXNextLayer, -0.5+\posY);    
    \draw [dashed] (\posX+0.15, -1.5+\posY) -- (\posX+\posXNextLayer, -0.5+\posY);

       \draw [dashed] (\posX+0.15, 0.5+\posY) -- (\posX+\posXNextLayer, 0.5+\posY);   
    \draw [dashed] (\posX+0.15, 0+\posY) -- (\posX+\posXNextLayer, 0.5+\posY);   
    \draw [dashed] (\posX+0.15, -0.5+\posY) -- (\posX+\posXNextLayer, 0.5+\posY);
    \draw [dashed] (\posX+0.15, -1.0+\posY) -- (\posX+\posXNextLayer, 0.5+\posY);    
    \draw [dashed] (\posX+0.15, -1.5+\posY) -- (\posX+\posXNextLayer, 0.5+\posY);

    \draw [dashed] (\posX+0.15, 0.5+\posY) -- (\posX+\posXNextLayer, -1.5+\posY);   
    \draw [dashed] (\posX+0.15, 0+\posY) -- (\posX+\posXNextLayer, -1.+\posY);   
    \draw [dashed] (\posX+0.15, -0.5+\posY) -- (\posX+\posXNextLayer, -1.5+\posY);
    \draw [dashed] (\posX+0.15, -1.0+\posY) -- (\posX+\posXNextLayer, -1.5+\posY);    
    \draw [dashed] (\posX+0.15, -1.5+\posY) -- (\posX+\posXNextLayer, -1.5+\posY);

  \def\sizeY{2.25}  \def\posX{4.75+\globalX}   \def\posY{+2.5+\globalY}   \def\depth{0.1}\def\perspectiveX{0.05}  \def\perspectiveY{\perspectiveX*1.5}  
  \draw (\posX,\posY) rectangle (\posX+\depth,\posY-\sizeY);     
  \draw (\posX+\perspectiveX, \posY+\perspectiveY) -- (\posX+\perspectiveX+\depth,  \posY+\perspectiveY);
 \draw (\posX+\perspectiveX+\depth,  \posY+\perspectiveY) -- (\posX+\perspectiveX+\depth,  \posY-\sizeY+\perspectiveY);  
    \draw (\posX+\perspectiveX,  \posY+\perspectiveY) -- (\posX,  \posY);    
   \draw (\posX+\perspectiveX+\depth,  \posY-\sizeY+\perspectiveY) -- (\posX+\depth,  \posY-\sizeY);      
   \draw (\posX+\perspectiveX+\depth,  \posY+\perspectiveY) -- (\posX+\depth,  \posY);

\end{scope}

\draw (6.8, 0) node {\small{{$      
 \left 
 .
\begin{bmatrix}
f_1 \\
f_2 \\
\vdots \\
f_t\\
\end{bmatrix}
\right.
$}}};  
    
\end{tikzpicture}
    \caption{The main task of DL-based iris feature extraction: given a dimensionless representation of the iris data, obtain its compact and representative representation - the feature set -  that is further used in the classification phase. }
        \label{fig:Representation}
    \end{center}
\end{figure}

 As illustrated in Fig.~\ref{fig:Representation}, the idea here is to analyze a dimensionless representation of the iris data and produce a feature vector that lies in a hyperspace (embedding) where recognition is carried out.

In this context, Boyd \emph{et} el.~\cite{DeepFeatExtr} explored five different sets of weights for the popular ResNet50 architecture to test if iris-specific feature extractors perform better than models trained for general tasks.  Minaee \emph{et} al.~\cite{VGGiris} studied the application of deep features extracted from VGG-Net for iris recognition, having authors observed that the resulting features can be well transferred to biometric recognition. Luo \emph{et} al.~\cite{9429938} described a DL model with spatial attention and channel attention mechanisms, that are directly inserted into the feature extraction module. Also, a co-attention mechanism adaptively fuses features to obtain representative iris-periocular features. Hafner \emph{et} al.~\cite{9415202} adapted the classical Daugman's pipeline, using convolutional neural networks to function as feature extractors. The DenseNet-201 architecture outperformed its competitors achieving state-of- the-art results both in the open and close world settings. Menotti \emph{et} al.~\cite{7029061} assessed how DL-based feature representations can be used in spoofing detection, observing that spoofing detection systems based on CNNs can be robust to attacks already known and  adapted, with little effort, to image-based attacks that are yet to come.

Yang \emph{et} al.~\cite{DualSANet} generated multi-level spatially corresponding feature representations by an encoder-decoder structure. Also, a spatial  attention feature fusion module was used to ensemble the resulting features more effectively. Chen \emph{et} al.~\cite{8995585} addressed the large-scale recognition problem and described an optimized center loss function (tight center) to attenuate the insufficient discriminating power of the cross-entropy function. Nguyen \emph{et} al.~\cite{OTS_CNN_Iris} explored the performance of state-of-the-art pre-trained CNNs on iris recognition, concluding that off-the-shelf CNN generic features are also extremely good at representing iris images, effectively extracting discriminative visual features and achieving promising results. Zhao \emph{et} al.~\cite{8689110} proposed a method based on the capsule network architecture, where a modified routing algorithm based on the dynamic routing between two capsule layers was described, with three pre-trained models (VGG16, InceptionV3, and ResNet50) extracting the primary iris features. Next, a convolution capsule replaces the full connection capsule to reduce the number of parameters. Wang and Kumar~\cite{IrisFCN_Dilated} introduced the concept of \emph{residual feature} for iris recognition. They described a residual network learning procedure with offline triplets selection and dilated convolutional kernels.

Other works have addressed the extraction of appropriate feature representations in multi-biometrics settings: Damer \emph{et} al.~\cite{9186011} propose to  jointly extract multi-biometric representations within a single DNN. Unlike previous solutions that create independent representations from each biometric modality, they create these representations from multi-modality (face and iris), multi-instance (iris left and right), and multi- presentation (two face samples), which can be seen as a fusion at the data level policy. Finally, concerned about the difficulty of performing reliable recognition in hand-held devices, Odinokikh \emph{et} al.~\cite{8987379} combined the advantages of handcrafted feature extractors and advanced deep learning techniques. The model utilizes shallow and deep feature representations in combination with characteristics describing the environment, to reduce the intra-subject variations expected in this kind of environments.

\subsection{Deep Learning-based Iris Matching Strategies}
The existing matching strategies can be categorized into three categories: (1) using conventional classifiers, such as SVM, RF, and Sparse Representation; (2) softmax-based losses; and (3) pairwise-based losses. A cohesive perspective of the most relevant recent DL-based
methods is given in Table 2, with the techniques appearing in chronographic (and then alphabetical)
order

\subsubsection{Conventional classifiers}

Various researchers have been using deep learning networks designed and pre-trained on the ImageNet dataset to extract iris feature representations, followed by a conventional classifier such as SVM, RF, Sparse Representation, etc. \cite{OTS_CNN_Iris,DeepFeatExtr,Boyd_Access_2020}. The key benefit of these approaches is the simplicity of ``plug and play'', where proven and pre-trained deep learning networks inherited from large-scale computer vision challenges are widely available and ready to be used \cite{OTS_CNN_Iris}. Another benefit is that there is no need for large scale iris image datasets to train these networks because they have already been trained on such large-scale datasets as ImageNet. Considering these networks usually contain hundreds of layers and millions of parameters, and require millions of images to train, using pre-trained networks is extremely beneficial.

\subsubsection{Iris Classification Networks}
Iris classification networks couple deep learning architectures with a family of softmax-based losses to classify an iris image into a list of known identities. Coupling a softmax loss with a backbone network enables training the backbone network in an end-to-end manner via popular optimization strategies such as back-propagation and steepest gradient decent. Compared to the conventional classifier approaches, the DL-based backbones in this category are learnable directly from the iris data, allowing them to better represent the iris. The key benefit is that it is similar to a generic image classification task, hence all designs and algorithms in the generic image classification task can be trivially applied with the iris image data. Typical examples of these iris classification networks are \cite{DeepIrisNet, DeepFeatExtr}. However, these softmax-based networks require the iris in the test image be known in the identity classes in the training set, which means the networks must be re-trained whenever a new class (\emph{i.e.} a new identity) is added. Gangwar \emph{et al.} proposed two backbone networks (\emph{i.e.} DeepIrisNet-A and DeepIrisNet-B) followed by a softmax loss for the iris recognition task \cite{DeepIrisNet}. Later, they proposed another backbone network, but still followed by a softmax loss to classify one normalized iris image into a pre-defined list of identity \cite{DeepIrisNet2}. 



\vspace{6px}
\noindent\textit{Backbone Network Architectures:} 
A wide range of backbone network architectures have been borrowed from generic image classification for the iris recognition task due to their similarity.
\begin{itemize}
    \item AlexNet: AlexNet is the most primitive and been shown as least accurate for iris recognition compared to others \cite{boyd2020iris,OTS_CNN_Iris}.
    \item VGG: Boyd \emph{et al.} \cite{boyd2020iris}, Nguyen \emph{et al.} \cite{OTS_CNN_Iris} and Minaee \emph{et al.} \cite{VGGiris} all experimented VGG16 .
    \item ResNet:  ResNet with its variants are the most popular backbone network architecture. Nguyen \emph{et al.} experimented ResNet152 \cite{OTS_CNN_Iris}. Boyd \emph{et al.} experimented three variants ResNet18, ResNet50 and ResNet152 in their post-mortem iris classification task \cite{boyd2020iris}.
    \item Inception: Zhao \emph{et al.} employed capsule network based on the InceptionV3 architecture \cite{8689110}. 
    \item EfficientNet: Hsiao \emph{et al.} \cite{9486782} employed EfficientNet to extract iris features. 
\end{itemize}


\subsubsection{Iris Similarity Networks}
Iris similarity networks couple deep learning architectures with a family of pairwise-based losses to learn a metric representing how similar or dissimilar two iris images are without knowing their identities. The pairwise loss aims to pull images of the same iris closer and push images of different irises away in the similarity distance space. Different to the iris classification networks which only operate in an identification mode on a pre-defined identity list, iris similarity networks operate across both verification and identification modes with an open set of identities \cite{IrisFCN}. Typical examples of these iris similarity networks are \cite{DeepIris,IrisFCN,IrisFCN_Dilated,ConstrainedIrisNet,9746090}. There are three key benefits of these networks: (i) verification and identification: iris similarity networks operate across both verification and identification modes; (ii) open set of identities: iris similarity networks operate on an open set of identities; and (iii) explicit reflection: iris similarity networks directly and explicitly reflect what we want to achieve, \emph{i.e.,} small distances between irises of the same subject and larger distances between irises of different subjects.

\vspace{6px}
\noindent\textit{Pairwise loss:} 
Nianfeng \emph{et al.} \cite{DeepIris} proposed a pairwise network, which accepts two input images and directly outputs a similarity score. They designed a pairwise layer which accepts two input images and encodes their features via a backbone network. The backbone network is trained iteratively to minimize the dissimilarity distance between genuine pairs (pairs of the same identity) and maximize the dissimilarity distance between impostor pairs (pairs of the different identities).

\vspace{6px}
\noindent\textit{Triplet loss:} 
Since the pairwise network is trained with separate genuine and impostor pairs, it may not converge well, which has been proven in the face recognition \cite{FaceNet}. Rather than using one pair of two images to update the training as in the pairwise loss for each training iteration, the triplet loss employs a triplet of three images: an anchor image, a positive image with the same identity and a negative image with a different identity \cite{FaceNet}. The backbone network is trained to simultaneously minimize the similarity distance between the positive and the anchor images and maximize the distance between the negative and the anchor images. Tailored for iris images, Zhao \emph{et al.} \cite{IrisFCN,IrisFCN_Dilated,IrisFCN_MaskRCNN} proposed Extended Triplet Loss (EPL) to incorporate a bit-shifting operation to deal with rotation in the normalized iris images. Nguyen \emph{et al.} also employed the ETL for their iris recognition network \cite{ConstrainedIrisNet,ComplexIrisNet}. Kuehlkamp \emph{et al.} \cite{Kuehlkamp_WACV_2022} proposed to improve the generic triplet loss function for iris recognition by forcing the distance to be positive (through the use of a sigmoid output layer), and adding a logarithmic penalty to the error. This modification allows the network to learn even when the difference between samples is negative and converge faster. Yan \emph{et al.} \cite{IFSR} extended the generic triplet loss to batch triplet loss, in which the triplet loss is calculated over a batch of $S$ subjects and $K$ images for each subject. Performing batch triplet loss is usually expected to have smooth loss function. Yang \emph{et al.} \cite{DualSANet} improved triplet selection method for training by Batch Hard \cite{BatchHard}.


\vspace{6px}
\noindent\textit{Backbone Network Architectures:} Different to the classification iris networks, similarity iris networks are usually designed with their own network architectures and are usually much ``shallower'' than the classification counterparts.

\begin{itemize}
    \item FCN: All similarity iris networks employ Fully Convolutional Networks (FCNs) instead of CNNs. Compared to CNNs, FCNs \cite{FCN} do not have fully connected layer, allowing the output map to preserve the original spatial information. This is important to iris recognition since the output map can preserve spatial correspondence with the original input image \cite{IrisFCN,ConstrainedIrisNet}, thus enabling pixel-to-pixel matching. Zhao \emph{et al.} \cite{IrisFCN} proposed a FCN architecture with 3 convolutional layers, followed by activation and pooling layers. Outputs of convolutional layers are up-sampled to the original input image size. The up-samples features are stacked and convolved by another convolutional layer to generate a 2-dimension features with the same size as the input image. Later, they extended the backbone network with dilated convolutions \cite{IrisFCN_Dilated}. Yan \emph{et al.} \cite{IFSR} employed a ResNet architecture and fine-tuned it with the triplet loss. Kuehlkamp \emph{et al.} only used a part of the ResNet architecture.


    \item NAS: Nguyen \emph{et al.} \cite{ConstrainedIrisNet} proposed to learn the network architecture directly from data rather than hand-designing it or using generic-image-classification architectures. They proposed a differential Neural Architecture Search (NAS) approach that models the architecture design process as a bi-level constrained optimization approach. This approach is not only able to search for the optimal network which achieves the best possible performance, but it can also impose constraints on resources such as model size or number of computational operations.

    \item Complex-valued: Observing that there is an intrinsic difference between the iris texture and generic object-based images where the iris texture is stochastic without consistent shapes, edges, or semantic structure, Nguyen \emph{et al.} \cite{ComplexIrisNet} argued the network architecture has to be better tailored to incorporate domain-specific knowledge in order to reach the full potential in the iris recognition setting. Another observation that they made is a majority of well-known handcrafted features such as IrisCode \cite{IrisCode} transformed iris texture image into a complex-valued representation first, then further encoded the complex-valued representation to get a final representation. They proposed to use fully complex-valued networks rather than popular real-valued networks. Complex-valued backbone networks better retain the phase, are more invariant to multi-scale, multi-resolution and multi-orientation, have solid correspondence with the classic Gabor wavelets \cite{ComplexCNN_Theory}, hence are much better suited to iris recognition than their real-valued counterparts. 
    
\end{itemize}

\begin{table*}
\centering
\caption{Cohesive comparison of the most relevant DL-based iris recognition methods (NIR: \emph{near-infrared}; VW: \emph{visible wavelength}). Methods are listed in chronological (and then alphabetical) order.}
\label{tab:Recognition}
\scriptsize
\begin{tabular}{|p{1.3cm}|p{1.6cm}|p{.38cm}|p{.32cm}|p{.32cm}|p{3.65cm}|p{4cm}|}
\hline
\multirow{2}{*}{\textbf{Category}}                                                 & \multirow{2}{*}{\textbf{Method}} & \multirow{2}{*}{\textbf{Year}} & \multicolumn{2}{c|}{\textbf{Data}}            & \multicolumn{1}{c|}{\multirow{2}{*}{\textbf{Datasets}}}                                                              & \multirow{2}{*}{\textbf{Features}}                                                                                                       \\ 
\cline{4-5}
                                                                                   &                                  &                                & \textbf{NIR}          & \textbf{VW}           & \multicolumn{1}{c|}{}                                                                                                &                                                                                                                                          \\ 
\hline
\multirow{6}{*}{\begin{tabular}[c]{@{}l@{}}Conventional\\classifiers\end{tabular}} & Menotti~\textit{et}~at. \cite{7029061}          & 2015                           &  \cmark                     &  \cmark                     & \textcolor[rgb]{0.2,0.2,0.2}{Biosec,~}\textcolor[rgb]{0.2,0.2,0.2}{LivDet-2013-Warsaw,~}\textcolor[rgb]{0.2,0.2,0.2}{MobBIOfake} & Shallow CNNs + SVM for~\textcolor[rgb]{0.2,0.2,0.2}{Spoofing Detection}                                                                  \\ 
\cline{2-7}
                                                                                   & Minaee~\textit{et}~al.~\cite{VGGiris}          & 2016                           &  \cmark                     &    \xmark                   & CASIA-Iris-Thousand, IITD                                                                                           & VGG + SVM                                                                                                                                \\ 
\cline{2-7}
                                                                                   & Nguyen \textit{et} al.~\cite{OTS_CNN_Iris}          & \multicolumn{1}{l|}{2017}      & \cmark & \xmark & \textcolor[rgb]{0.2,0.2,0.2}{ND-CrossSensor-2013, CASIA-Iris-Thousand}                                                & \textcolor[rgb]{0.2,0.2,0.2}{AlexNet, VGG, Google Inception, ResNet, DenseNet + SVM}                                                     \\ 
\cline{2-7}
                                                                                   & Boyd \textit{et} al.~\cite{DeepFeatExtr}            & 2019                           &  \cmark                     & \cmark                      & CASIA-Irisv4-Interval, IITD, UBIRIS.v2                                                                                   & ResNet50 + SVM                                                                                                                          \\ 
\cline{2-7}
                                                                                   & Boyd \textit{et} al.~\cite{Boyd_Access_2020}            & 2020                           &  \cmark                     &  \cmark                     & \textcolor[rgb]{0.2,0.2,0.2}{DCMEO1,~}\textcolor[rgb]{0.2,0.2,0.2}{Warsaw}                                 & AlexNet, ResNet, VGG, DenseNet + Cosine, Euclidean, MSE                                                                                  \\ 
\cline{2-7}
                                                                                   & Hafner \textit{et} al.~\cite{9415202}          & 2021                           &  \cmark                     &  \xmark                     & \textcolor[rgb]{0.2,0.2,0.2}{CASIA-Iris-Thousand}                                                                    & \textcolor[rgb]{0.2,0.2,0.2}{ResNet101 +~}\textcolor[rgb]{0.2,0.2,0.2}{DenseNet-201 +~}\textcolor[rgb]{0.2,0.2,0.2}{Cosine Similarity}  \\ 
\hhline{|=======|}
\multirow{6}{*}{\begin{tabular}[c]{@{}l@{}}Classification\\Networks\end{tabular}}  & Gangwar~\textit{et} al.~\cite{DeepIrisNet}         & 2016                           &  \cmark                     & \xmark                      & ND-IRIS-0405, ND-CrossSensor-2013  & DeepIrisNet                                                                                                                              \\ 
\cline{2-7}
                                                                                   & Gangwar \textit{et} al.~\cite{DeepIrisNet2}         & 2019                           &  \cmark         & \cmark                      & ND-IRIS-0405, UBIRIS.v2, MICHE-I, CASIA-Irisv4-Interval                                                              & DeepIrisNetV2                                                                                                                            \\ 
\cline{2-7}
                                                                                   & Odinokikh \textit{et} al.~\cite{8987379}                & \multicolumn{1}{l|}{2019}      & \cmark & \xmark & CASIA-Iris-M1-S2, CASIA-Iris-M1-S3, Iris-Mobile                                                                       & Feature Fusion + Softmax                                                                                                                 \\ 
\cline{2-7}
                                                                                   & Zhao \textit{et} al.\cite{8689110}             & \multicolumn{1}{l|}{2019}      & \cmark & \xmark & \textcolor[rgb]{0.2,0.2,0.2}{JluIrisV3.1, JluIrisV4, CASIA-Irisv4-Lamp}                                                  & Capsule network + Softmax                                                                                                                \\ 
\cline{2-7}
                                                                                   & Chen \textit{et} al. \cite{8995585}            & \multicolumn{1}{l|}{2020}      & \cmark & \xmark & \textcolor[rgb]{0.2,0.2,0.2}{ND-IRIS-0405, CASIA-Iris-Thousand, IITD cross sensor}                                   & \textcolor[rgb]{0.2,0.2,0.2}{T-Center loss~}                                                                                             \\ 
\cline{2-7}
                                                                                   & Luo \textit{et} al.\cite{9429938}              & 2021                           & \cmark                      &  \xmark                     & \textcolor[rgb]{0.2,0.2,0.2}{ND-IRIS-0405,~}\textcolor[rgb]{0.2,0.2,0.2}{CASIA-Iris-Thousand}                        & Attention + Softmax Loss + Center Loss                                                                                                   \\ 
\hhline{|=======|}
\multirow{10}{*}{\begin{tabular}[c]{@{}l@{}}Similarity\\Networks\end{tabular}}     & Nianfeng \textit{et} al.~\cite{DeepIris}        & 2016                           & \cmark                      &  \xmark                     & \textcolor[rgb]{0.18,0.18,0.18}{Q-FIRE, CASIA-Cross-Sensor~}                                                         & DeepIris                                                                                                                                 \\ 
\cline{2-7}
                                                                                   & Zhao \textit{et} al.~\cite{IrisFCN}            & 2017                           &  \cmark                     &  \cmark                     & CASIA-Irisv4-Interval, IITD, UBIRIS.v2                                                                               & UniNet (FeatNet+MaskNet) +~Extended Triplet Loss                                                                                         \\ 
\cline{2-7}
                                                                                   & Damer \textit{et} al.~\cite{9186011}                     & \multicolumn{1}{l|}{2019}      & \cmark & \xmark & Biosecure, CASIA-Iris-Thousand/Lamp/Interval                                                                         & Inception + Triplet Loss                                                                                                                 \\ 
\cline{2-7}
                                                                                   & Wang \textit{et} al.~\cite{IrisFCN_Dilated}            & 2019                           &   \cmark                    &   \cmark                    & CASIA-Irisv4-Interval, IITD, UBIRIS.v2                                                                               & FeatNet + Dilated Convolution +~Extended Triplet Loss                                                                                    \\ 
\cline{2-7}
                                                                                   & Zhao \textit{et} al.~\cite{IrisFCN_MaskRCNN}                        & 2019                           &  \cmark                     &  \xmark                     & ND-Iris-0405, Casia-Irisv4-Distance, IITD 
                                                                                   & FeatNet + Mask RCNN +~Extended Triplet Loss                                                                                              \\ 
\cline{2-7}
                                                                                   & Nguyen~\textit{et} al.~\cite{ConstrainedIrisNet}          & 2020                           &  \cmark                     & \cmark                      & CASIA-v4-Distance, UBIRIS.v2, ND-CrossSensor-2013                                                                               & Constrained Design Backbone +~Extended Triplet Loss                                                                                      \\ 
\cline{2-7}
                                                                                   & Yan \textit{et} al.~ \cite{IFSR}            & 2021                           &  \cmark                     &  \xmark                     & CASIA-Iris-Thousand                                                                                                  & Spatial Feature Reconstruction + Triplet Loss                                                                                            \\ 
\cline{2-7}
                                                                                   & Yang \textit{et} al.~\cite{DualSANet}             & \multicolumn{1}{l|}{2021}      & \cmark & \xmark & CASIA-Irisv4-Thousand, CASIA-Irisv4-Distance, IITD                                                                  & Dual Spatial Attention Network +~Batch Hard                                                                                              \\ 
\cline{2-7}
                                                                                   & Nguyen~\textit{et} al.~\cite{ComplexIrisNet}          & 2022                           &  \cmark                     & \cmark                      & ND-CrossSensor-2013, CASIA-Iris-Thousand, UBIRIS.v2                                                                  & Complex-valued Backbone +~Extended Triplet Loss                                                                                          \\ 
\cline{2-7}
                                                                                   & Kuehlkamp \textit{et} al. \cite{Kuehlkamp_WACV_2022}        & 2022                           &  \cmark                     &    \cmark                   & 
                                                    DCMEO1, DCMEO2, Warsaw-Post-Mortem v2.0 & ResNet + Triplet Loss                                                                                                                    \\
\hline
\end{tabular}
\end{table*}

\subsection{End-to-end Joint Iris Segmentation+Recognition Networks}
Almost all existing approaches perform segmentation and normalization to transform an input image to a normalized rectangular 2D representation before recognition as this simplifies the representation learning. As segmentation and recognition may require a separate network themselves, this would cause redundancy in both computation and training, further slowing down an DL-based iris recognition approach. Several researchers have looked at approaches to perform end-to-end networks. One category is to perform segmentation-less recognition. Another category is to jointly learn segmentation and recognition using an unified network via multi-task learning.

\vspace{6px}
\noindent\textit{Segmentation-less:}
These approaches feed the cropped iris images directly into a deep learning network to extract features. For example, Kuehlkamp \emph{et al.} \cite{Kuehlkamp_WACV_2022} used Mask R-CNN for semantic segmentation and fed the cropped iris region directly into a ResNet50 to extract features. Similarly, Chen \emph{et al.} \cite{E2Eiris} also fed the cropped iris images directly into a DenseNet. Rather than feeding the cropped iris images directly, Proenca \emph{et al.} transformed the cropped region (which is detected by SSD) into a polar representation first, then fed the polar representation into the VGG19 for extracting features \cite{Segmentation-Less}.

\vspace{6px}    
\noindent\textit{Multi-task:}
Segmentation and recognition can be jointly learned with one unified network. This paves a way for multi-task learning. However, segmentation and recognition may require different number of layers, hence research is required to perform using different intermediate layers for each task. To the best knowledge, there does not exist any approach to explore this direction.

\section{Deep Learning-Based Iris Presentation Attack Detection}
\label{sec:presentationattack}
%

In parallel to the popularity of biometrics, the security of these systems against attacks has become of paramount importance.
The most common attack
is a \textit{Presentation Attack} (PA), which refers to presenting a fake sample to the sensor. 
The goal can be either to \textit{impersonate} somebody else identity (also known as \textit{Impostor Attack Presentation}), or to \textit{conceal} the own identity (also known as \textit{Concealer Attack Presentation}).
Via impostor attacks, a person could also enroll fraudulently, allowing a continuous manipulation of the system.
The previous acronyms and terms in italics correspond to the vocabulary recommended in the series of ISO/IEC 30107 standards of the ISO/IEC Subcommittee 37 (SC37) on Biometrics \cite{ISO/IEC30107-3}, which we will follow in the rest of this section. 
\textit{Presentation Attack Instruments} (PAI) used to carry out impostor attacks are typically generated from \textit{bona fide} images of an iris 
from an individual who has legitimate access to the system. The iris is printed on a piece of paper (printout attack) or displayed on a screen (replay attack) and then presented to the sensor.
%
The iris of deceased individuals can also be used as PAI, since the texture remains intact for some hours \cite{Trokielewicz_18_btas}.
Theoretically, it would be possible to print a genuine iris texture into a contact lens as well, 
although this has not been successfully demonstrated yet \cite{boyd2020iris}.
Concealer attacks, on the other hand, are commonly done via textured contact lenses that obscure 
or alter properties of the eye (such as color) to prevent the system from identifying the user.  
%
Synthetic iris images \cite{Yadav19CVPRW} not belonging to any specific identity could be used for similar purposes.
%
Concealers can also present their legitimate iris, but in a way not expected by the system, e.g. closing eyelids as much as possible, looking to the sides (off-axis gaze), rotating the head, etc.

Two challenges of PAs is that they happen outside the physical limits of the system, and they do not require specific knowledge of its inner workings, 
or any technical knowledge at all.
Thus, if no properly tackled, they can derail public perception of even the most reliable biometric modality. 
It is even more critical if authentication is done without any supervision.
\textit{Presentation Attack Detection} (PAD) methods to counteract such attacks can be done \cite{Galbally_16_iwbf}: $i$) at the \textit{hardware} (or sensor) level, using additional illuminators or sensors that detect intrinsic properties of a living eye or responses to external stimuli (like pupil contraction or reflection), 
or $ii$) at the \textit{software} level, using only the footprint of the PA (if any) left in the same images captured with the standard sensor that will be employed for authentication.
%
%
%
%
Software-based techniques are in principle less expensive and intrusive, since they do not demand extra hardware, and they will be the focus of this section.

Two comprehensive surveys on PAD are \cite{Czajka_ACM_2018} (2018) and \cite{boyd2020iris} (2020). 
While DL techniques were residual in the 2018 survey, they rose in popularity thereafter.
We build this section upon the latest survey and summarize the most important developments in DL-PAD since it was published (Table~\ref{tab:PAD}). A descriptive summary of the datasets employed is given later in Section~\ref{sec:opensources}.
The aim of PAD is to classify an image either as a \textit{bona fide} or an \textit{attack} presentation, so it is usually modeled as a two-class classification task.
Typical strategies mimic the trend of the previous section when applying DL to iris recognition:
either a CNN backbone is used to extract features that will feed a conventional classifier, or the network is trained end-to-end to do the classification itself.
Some hybrid methods 
also combine traditional hand-crafted with deep-learned features.
In the same manner, the network may be initialized e.g. on the ImageNet dataset to take advantage of such large generic corpus, since available iris PAD data is more scarce.  
%
%
%
%
Another strategy also employed widely in the PAD literature is to use adversarial networks, where a GAN \cite{10.5555/2969033.2969125} is trained to generate synthetic iris images that the discriminator must use to detect attack samples. 

\subsection{CNNs for Feature Extraction}
Since each layer of a CNN represents a different level of abstraction, 
Fang \textit{et al}. \cite{Fang20FUSION} 
fused the features from the last four convolutional layers of two models (VGG16, MobileNetv3-small). 
The features are projected to a lower dimensional space by PCA and either concatenated for classification with SVM (feature fusion) or the classification scores of each level combined (score fusion). 
%
Using two databases of printouts and textured contact lenses, the method showed superiority over the use of the different layers individually, or the feature vector from the next-to-last layer of the networks.

\subsection{End-to-end Classification Networks}

Arora and Bathia \cite{Arora20IJSAEM} trained a 
CNN with 10 convolutional layers to detect contact lenses and printouts. 
Rather than using the entire image, the network is trained on patches from all parts of the iris image.
The system showed superior performance compared to state-of-the-art methods which at that time, according to the paper, were mostly based on hand-crafted features.

Focusing on embedded low-power devices, Peng \textit{et al}. \cite{Peng20aeeca} adopted a Lite Anti-attack Iris Location Network (LAILNet) based on three dense blocks featuring depthwise separable convolutions to reduce the number of parameters.
The algorithm demonstrated very good performance on three databases with printouts, synthetic irises, contact lenses and artificial plastic eyes.

Also focusing on mobiles, Fang \textit{et al}. \cite{Fang21IVC,Fang20ijcb} used MobileNetv3-small.
The contribution lies in the division of the normalized iris image into 
overlapped micro-stripes which are fed individually, and a decision reached by majority voting.
The claimed advantages are that the classifier is forced to focus on the iris/sclera boundaries (given by their exact micro-stripes), the input dimensionality is lower and the amount of samples is higher (reducing overfitting), and the impact of imprecise segmentation is alleviated.
Using three databases with contact lenses and printouts, the paper featured an extensive experimentation with cross-database, cross-sensor, and cross-attack setting. 

Sharma and Ross \cite{Sharma20ijcb} proposed D-NetPAD, based on  DenseNet121, 
chosen due to benefits such as maximum flow of information given by dense connections to all subsequent layers,
or fewer parameters compared to counterparts like ResNet or VGG.
The PAI included printouts, artificial eye, cosmetic contacts, kindle replay, and transparent dome on print, 
with experiments substantiating the effectiveness of the method on cross-PAI, cross-sensor and cross-database scenarios.

Chen and Ross \cite{Chen21wacv} proposed an explainable
attention-guided 
detector (AG-PAD).
To do so, the feature maps of a DenseNet121 
were fed into two modules 
that independently capture inter-channel and inter-spatial feature dependencies.
The outputs were then fused via element-wise sum to capture complementary attention features from both channel and spatial dimensions.
With three datasets containing colored contact lenses, artificial eyes (Van Dyke/Doll fake eyes), printouts, and textured contact lenses, 
the attention modules are shown to improve accuracy over the baseline network.
Using heatmap visualization, it is also shown that the attention modules force the network to attend to the annular iris textural region which, intuitively, plays a vital role for PAD.

Spatial attention was also explored by Fang \textit{et al}. \cite{Fang21IJCB}.
%
%
To find local regions that contribute the most to make accurate decisions and capture pixel/patch-level cues, they proposed an attention-based pixel-wise binary supervision (A-PBS) method. To capture different levels of abstraction, they perform multi-scale fusion by adding spatial attention modules to feature maps from three levels of a DenseNet backbone.
%
%
%
Using six datasets with textured lenses and printouts, they outperformed previous state-of-the-art including scenarios with unknown attacks, sensors, and databases.

Given the difficulty of collecting iris PAD data, most databases contain, at most, a few hundred subjects.
%
%
%
To address this, Fang \textit{et al}. \cite{Fang21MVA} studied data augmentation techniques that modify position, scale or illumination. 
Using three architectures (ResNet50, VGG16, MobileNetv3-small) and three databases with printouts and textured contact lenses, they found that data augmentation improves PAD performance significantly, but each technique has a positive role on a particular dataset or CNN. 
They also explored the selection of augmentation techniques, 
finding, again, no consensus regarding the best combination,
which was attributed to differences in capture environment, subject population, scale of the different datasets or imbalance between bona fide and attack samples.

Gupta \textit{et al}. \cite{Gupta20ICPR} proposed 
MVANet, with 5 convolutional layers and 3 branches of fully connected layers. They addressed the challenge of unseen databases, sensors, and imaging environment on textured contact lenses detection.
%
%
The size of each layer of MVANet is different, thus capturing different features.
They used three databases, each one captured in different settings (indoor/outdoor, different times of the day, varying weather, fixed/mobile sensors, etc.), with MVANET trained in one database at a time and tested on the other two.
As baseline, they fine-tuned 
three popular CNNs (VGG16, ResNet18, DenseNet) initialized on ImageNet.
The proposed network is shown to perform consistently better and more uniformly on the 
test databases than the baseline approaches. 

Sharma and Ross \cite{Sharma20ICPR} studied the viability of Optical Coherence Tomography (OCT). 
OCT provides a cross-sectional view of the eye, whereas traditional NIR or VW imaging provides 2D textural data.
The PAIs considered are artificial eyes (Van Dyke eyes) and cosmetic
lenses, evaluated on three different CNNs (VGG19, ResNet50, DenseNet121). 
By both intra- (known PAs) and cross-attack (unknown PAs) scenarios, OCT is determined as a viable solution, although hardware cost is still a limiting factor. Indeed, OCT outperforms NIR and VW in the intra-attack scenario, while NIR generalizes better to unseen PAs. 
Cosmetic lenses also appear to be more difficult to detect than artificial eyes with any modality.
Via heatmaps, it is seen as well that the fixation regions are different for each imaging modality and for each PAI, which could be a source of complementarity.

Zhang \textit{et al}. \cite{Zhang20ICPR} proposed a Weighted Region Network (WRN) to detect
cosmetic lenses that includes a local attention Weight Network (for evaluating
the discriminating information of different regions) and a global classification Region Network (for characterizing global features). 
Such strategy considers both the entire image and the attention effect by assigning different
weights to regions.
The mentioned networks are applied to a VGG16 backbone.
%
%
The reported results showed improved performance compared to the state-of-the-art over three different databases.

The works by Agarwal \textit{et al}. \cite{Agarwal22TBIOM,AGARWAL22PRL} evaluated the detection of contact lenses.
In \cite{Agarwal22TBIOM}, they trained a siamese CNN of 5 convolutional layers on two different inputs (the original image and its CLAHE version), which are then combined by weighted score fusion of the softmax layer.
Adding a processed version of the raw image attempts to enhance the feature extraction capabilities of the CNN.
A similar strategy is followed in \cite{AGARWAL22PRL}, but here they used a siamese contraction-expansion CNN, and the processed image is a edge-enhanced image obtained via Histogram of Oriented Gradients (HOG).
Another difference was the use of feature-level fusion of the next-to-last CNN feature vectors, testing different strategies (vector addition, multiplication, concatenation and distance). 
The papers employed several databases, with an extensive protocol including unseen subjects, 
environments (indoor vs outdoor) and 
databases (sensors) that showcases the strength of the solutions against cross-domain changes. 
The methods also showed superiority against popular CNN models (VGG16, ResNet18, DenseNet) and the popular LBP and HOG hand-crafted features.

Gautam \textit{et al}. \cite{Gautam22DSP} proposed a Deep Supervised Class Encoding (DSCE) approach consisting of an Autoencoder that exploits class information, and minimizes simultaneously the reconstruction and classification errors during training.
Three datasets were used, containing textured lenses, printouts and synthetic images, showing superiority over a variety of hand-crafted and deep-learned features.
%

Tapia \textit{et al}.\cite{Tapia22TIFS} used a two-stages serial architecture based on a modified MobiletNetv2. 
A first network was trained to only distinguish two classes (bona fide vs attack).
If it votes \textit{bona fide}, the image is sent to a second network trained to classify it among three or four classes (bona fide or a different type of PAI: contact lenses, printout, or cadaver).
Four databases were combined to obtain a super-set with the different PAIs, and class-weights were also incorporated into the loss to compensate imbalance. 
The paper applied contrast enhancement (CLAHE), and an aggressive data augmentation 
(rotation, blurring, contrast change, Gaussian noise, edge enhancement, image region dropout, etc.).
They tested two image sizes, 224$\times$224 and 448$\times$448, observing that the extra detail of a higher resolution image results in more effective features. 
%
%
%
The paper also carried out leave-one-out PAI tests for open-set evaluation, 
showing robustness in detecting unknown attacks.

\subsection{Hybrid Methods}

Choudhary \textit{et al}. \cite{Choudhary22JAIHC,Choudhary22TIFS} applied a Friedman test-based 
selection method to identify the best features of a set of hand-crafted and deep-learned ones.
Each feature method feeds a SVM classifier, and the scores of the individual SVMs are fused via weighted sum.
A preliminary version of \cite{Choudhary22TIFS} without feature selection appeared in \cite{Choudhary21ICCE}.
The databases of \cite{Choudhary22JAIHC} include a medley of different PA (printouts, synthetic irises, artificial eyeballs, etc.), although the feature selection and classification methods are trained and evaluated separately on each database.
The authors observed a saturation 
after a certain number of features are combined, and a superiority of the score-level fusion over other methods such as majority voting, feature-level fusion, and rank-level fusion.
The work \cite{Choudhary22TIFS}, on the other hand, concentrated on textured contact lenses attack, with 
an extensive set of evaluations including single sensor, cross-sensor and combined sensor experiments.
Apart from the generic live vs attack scenario, it also reports binary and ternary classification across the different types of real (normal iris, soft lens) and fake (textured) classes. 
Naturally, the cross-sensor error is larger compared to single-sensor, and the combined sensor error is also observed to be slightly larger.
The latter is attributed to the larger intraclass variation created when images from different sensors are combined.
In any case, an improvement of performance over previous works with the three datasets employed is observed after the proposed feature selection and score-level fusion method.

\subsection{Adversarial Networks}
\label{sec:PAD-adversarial}

Generative methods have been used by some approaches, either to use the trained discriminator for iris PAD,
or to generate synthetic samples and augment under-represented classes. 
In this direction, Yadav and Ross \cite{Yadav21WACV} proposed CIT-GAN (Cyclic Image Translation Generative Adversarial Network) for multi-domain style transfer to generate synthetic samples 
of several PAIs (cosmetic contact lenses, printed eyes, artificial eyes and kindle-display attack).
To do so, image translation is driven by a Styling Network that learns style characteristics of each given domain.
It also employs a Convolutional Autoencoder in the generator for image-to-image style translation, which takes a domain label as input along with an image.
This is different than previous works of the same authors \cite{Yadav20WACV,Yadav19CVPRW} which employed the traditional generator/discriminator approach driven by a noise vector. 
%
%
Different PAD methods 
using hand-crafted (BSIF, DESIST) and deep features (VGG16, D-NetPAD, AlexNet) were evaluated, demonstrating that 
they can be improved by adding synthetically generated data.
The quality of synthetic images is also superior to a competing generative method (Star-GAN v2), measured via FID score distributions.

\subsection{Open Research Questions in Iris PAD}

One of the open research issues is to design robust iris PAD methods with cross-sensor and cross-database capabilities, so they 
generalize to unseen imaging conditions.
Attackers are constantly developing new attack methodologies to circumvent PAD systems, so an even more important issue is 
unseen PAIs (i.e. cross-PAI capabilities) \cite{Sharma21RIC}.
Great results have been achieved on detecting known attack types (known as \textit{closed-set} recognition), although cross-database evaluation (training in one database an testing in others) still appears as a difficult challenge due to changes in sensors, acquisition environments, or subjects.
Moreover, generalizing to attacks that are unknown at the time of training (\textit{open-set} recognition) is even a greater challenge for state-of-the-art methods \cite{Fang21IVC}.
%
%
%
%
Part of the problem lies into the limited size of existing databases, which is an issue for data-hungry DL approaches.
Some solutions, as studied by some of the methods above, are data augmentation by geometric or illumination modifications \cite{Fang21MVA}, or creating additional synthetic data 
via generative methods \cite{Yadav21WACV}.
Human-aided DL training is another promising avenue. Indeed, humans and machines cooperating in vision tasks is not new, and this strategy is finding its way into DL as well \cite{Boyd_CYBORG_2021,Boyd_WACV_2022}.
For example, Boyd \textit{et al}. \cite{Boyd_WACV_2022} analyzed the utility of human judgement about salient regions of images to improve generalization of DL models.
Asked about regions that humans deem important for their decision about an image, the work proposed to transform the training data to incorporate such opinions, demonstrating an improvement in accuracy and generalization in leave-one-attack-type-out scenarios.  
%
%
%
In a similar work, Boyd \textit{et al}. \cite{Boyd_CYBORG_2021} incorporated annotated saliency maps into the loss function to penalize large differences with human judgement.

Recently, concerns have emerged about the observed bias of DL methods that leads to discriminatory performance differences based on the user´s demographics, with face biometrics being the most talked-about and many companies and authorities banning its use \cite{Jain21tbiom}.
%
%
%
Obviously, this issue appears in iris PAD as well, as addressed by Fang \textit{et al}. \cite{Fang20eusipco}.
Using three baselines based on hand-crafted and DL approaches 
and a database of contact lenses, the authors showed  a significant difference in the performance between male and female samples. 
In dealing with this phenomenon, examination of biases towards eye color or race are another directions worthwhile to consider.

Some elements considered as PAIs in this section, such as cosmetic lenses, may be worn normally by users without the purpose of fooling the biometric system,
%
as it is the case of facial retouching via make-up, digital beautification or augmented reality \cite{DBLP:journals/corr/abs-2110-08934}.
This poses the question of whether it is possible to use such images for authentication, while diminishing the effect in the recognition performance. 
Suggested alternatives have been to detect and match portions of live iris tissue still visible \cite{Parzianello22WACVW} or incorporate ocular information of the surrounding area \cite{[Alonso16]}.
%
%
%
%
Unfortunately, in iris biometrics, recognition with textured contact lenses remains a hard problem to solve.

Another under-researched task is iris PAD in the visible spectrum. The majority of studies and datasets (Section~\ref{sec:opensources}) employ near-infrared illumination and specific iris close-up sensors. However, in some environments such as mobile or distant capture, such sensing is not guaranteed \cite{Nigam15IF}.

\begin{table*}[h]
    \centering
     \caption{Cohesive comparison of the most relevant DL-based iris Presentation Attack Detection methods after the surveys \cite{Czajka_ACM_2018,boyd2020iris} (NIR: \emph{near-infrared}; VW: \emph{visible wavelength}). Methods are listed in chronological (and then alphabetical) order.}

    \scriptsize
\begin{tabular}{|p{1.3cm}|p{1.6cm}|p{.38cm}|p{.32cm}|p{.32cm}|p{3.65cm}|p{4cm}|}
\hline
\multirow{2}{*}{\textbf{Category}}                                                 & \multirow{2}{*}{\textbf{Method}} & \multirow{2}{*}{\textbf{Year}} & \multicolumn{2}{c|}{\textbf{Data}}            & \multicolumn{1}{c|}{\multirow{2}{*}{\textbf{Datasets}}}                                                              & \multirow{2}{*}{\textbf{Features}}                                                                                                       \\ 
\cline{4-5}
                                                                                   &                                  &                                & \textbf{NIR}          & \textbf{VW}           & \multicolumn{1}{c|}{}                                                                                                &                                                                                                                                          \\ 
\hline

\multirow{1}{*}{\begin{tabular}[c]{@{}l@{}}Feature\\Extraction\end{tabular}}   

  & Fang \textit{et al}. \cite{Fang20FUSION}  & 2020 & \cmark & \xmark & LivDet-2017 (IIITD-WVU, ND-
CLD) & VGG16, MobileNetv3-small (multi-layer features) + PCA + SVM    \\ 
\cline{2-7}

\hhline{|=======|}

\multirow{1}{*}{\begin{tabular}[c]{@{}l@{}}End-to-end\\Training\end{tabular}}   

   & Arora and Bathia \cite{Arora20IJSAEM}  & 2020 & \cmark & \xmark & LivDet-2017 (IIITD-WVU) & CNN with patch input    \\ 
\cline{2-7}

   &  Peng \textit{et al}. \cite{Peng20aeeca}  & 2020 & \cmark & \xmark & IPITRT, CASIA-Iris-v4, CASIA-Iris-Fake & LAILNet lightweight CNN    \\ 
\cline{2-7}

 & Sharma and Ross \cite{Sharma20ijcb}  & 2020 & \cmark & \xmark & Proprietary, LivDet-2017 (IIITD-WVU, ND-CLD, Warsaw, Clarkson) & DenseNet121 pre-trained on ImageNet   \\ 
\cline{2-7}

   & Chen and Ross \cite{Chen21wacv} & 2021 & \cmark & \xmark & JHU-APL, LivDet-2017 (Warsaw, ND-CLD)  & DenseNet121 pre-trained on ImageNet + AG-PAD channel and spatial attention    \\ 
\cline{2-7}

   &  Fang \textit{et al}. \cite{Fang21IVC}  & 2021 & \cmark & \xmark & LivDet-2017 (IIITD-WVU, ND-CLD), ND-CLD-15,  & MobileNetv3-small with micro-stripes    \\ 
\cline{2-7}

  &  Fang \textit{et al}. \cite{Fang21IJCB} & 2021 & \cmark & \xmark & LivDet-2017 (IIITD-WVU, ND-CLD, Clarkson), ND-CLD-13, ND-CLD-15, IIITD-CLI    & DenseNet + A-PBS spatial attention    \\ 
\cline{2-7}

   & Fang \textit{et al}. \cite{Fang21MVA}  & 2021 & \cmark & \xmark & LivDet-2017 (IIITD-WVU, ND-CLD, Clarkson) & ResNet50, VGG16, MobileNetv3-small    \\ 
\cline{2-7}

   & Gupta \textit{et al}. \cite{Gupta20ICPR} & 2021 & \cmark & \xmark & MUIPA, UnMIPA, IIITD-CLI    & CNN with multi-branch classification    \\ 
\cline{2-7}

   & Sharma and Ross \cite{Sharma20ICPR} & 2021 & \cmark & \cmark & OCT, NIR and VW images & VGG19, ResNet50, DenseNet121    \\ 
\cline{2-7}

   & Zhang \textit{et al}. \cite{Zhang20ICPR}  & 2021 & \cmark & \xmark & ND-CLD-13, CASIA-Iris-Fake, IF-VE & VGG16 + WRN local attention and global classification    \\ 
\cline{2-7}

   & Agarwal \textit{et al}. \cite{AGARWAL22PRL} & 2022 & \cmark & \xmark & MUIPA, UnMIPA, IIITD-CLI, LivDet-2017 (IIITD-WVU), ND-PSID & Siamese contraction-expansion CNN, feature fusion   \\ 
\cline{2-7}

&  Agarwal \textit{et al}. \cite{Agarwal22TBIOM} &  2022 & \cmark & \xmark & MUIPA, UnMIPA, IIITD-CLI, LivDet-2017 (IIITD-WVU), ND-PSID, NDIris3D & Siamese CNN, score fusion    \\ 
\cline{2-7}


 & Gautam \textit{et al}. \cite{Gautam22DSP} & 2022 & \cmark & \xmark & SYN, IIITD-CLI, IIITD-IS & Autoencoder with reconstruction and classification loss \\ 
\cline{2-7}



& Tapia \textit{et al}. \cite{Tapia22TIFS} & 2022 & \cmark & \cmark & LivDet-2020, Iris-CL1, Warsaw-Post-Mortem v3.0 & MobileNetv2, data augmentation, class-weights    \\ 
\cline{2-7}

\hhline{|=======|}

\multirow{1}{*}{\begin{tabular}[c]{@{}l@{}}Hybrid\\Methods\end{tabular}}  

   &      Choudhary \textit{et al}. \cite{Choudhary22JAIHC}        & 2022 & \cmark & \xmark & IIITD-CLI, ND-CLD-13, CASIA, LivDet-2017 (IIITD-WVU, ND-CLD, Clarkson)  & MBISF (domain-specific filters), SIFT, Haralick, DenseNet, VGG8 + SVM classification     \\ 
\cline{2-7}

   & Choudhary \textit{et al}. \cite{Choudhary22TIFS}   &  2022 & \cmark & \xmark & IIITD-CLI, ND-CLD-13, LivDet-2017 (Clarkson) & MBSIF (generic filters),
MBSIF (domain-specific filters), SIFT, LBPV, DAISY, DenseNet121 + SVM classification    \\ 
\cline{2-7}

\hhline{|=======|}

\multirow{1}{*}{\begin{tabular}[c]{@{}l@{}}Adversarial\\Networks\end{tabular}}  

   & Yadav and Ross \cite{Yadav21WACV} & 2021 & \cmark & \xmark & Casia-Iris-Fake, Berc-iris-fake, ND-CLD-15, LivDet-2017, MSU-IrisPA-01 & BSIF, DESIST, VGG16, D-NetPAD, AlexNet    \\ 
\cline{2-7}

\hline
\end{tabular}

    \label{tab:PAD}
\end{table*}

\section{Deep Learning-Based Forensic Iris Recognition}
\label{sec:forensic}

Iris recognition has become the next biometric mode (in addition to face, fingerprints and palmprints) considered for large-scale forensic applications \cite{FBI_NGI}, and coincides in time with discoveries made in recent years about possibility to employ iris in recognition of deceased subjects. This includes both matching of iris patterns acquired a few hours after death with those with longer PMIs (Post-Mortem Intervals), ranging from days \cite{Sauerwein_JFO_2017,Bolme_BTAS_2016,Trokielewicz_ICB_2016,Trokielewicz_BTAS_2016} to several weeks after demise \cite{Trokielewicz_TIFS_2019,Boyd_Access_2020}, as well as matching patterns acquired before death with those collected post-mortem \cite{Sansola_MastersThesis_2015}. 

\begin{figure}[!htb]
     \begin{subfigure}{0.3\textwidth}
         \centering
         \includegraphics[width=\textwidth]{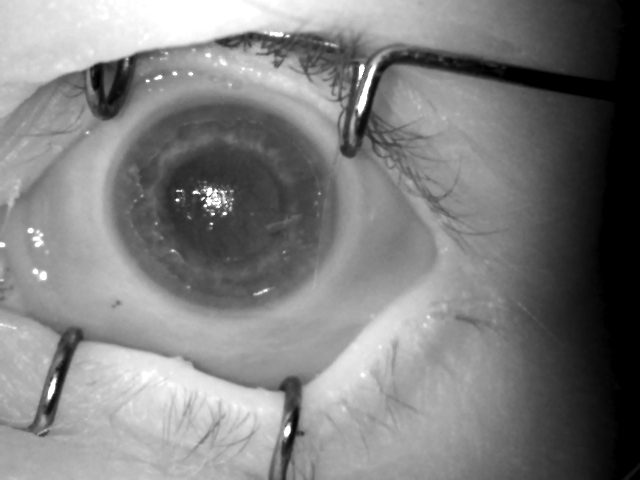}
         \caption{}
     \end{subfigure}\hfill
     \begin{subfigure}{0.3\textwidth}
         \centering
         \includegraphics[width=\textwidth]{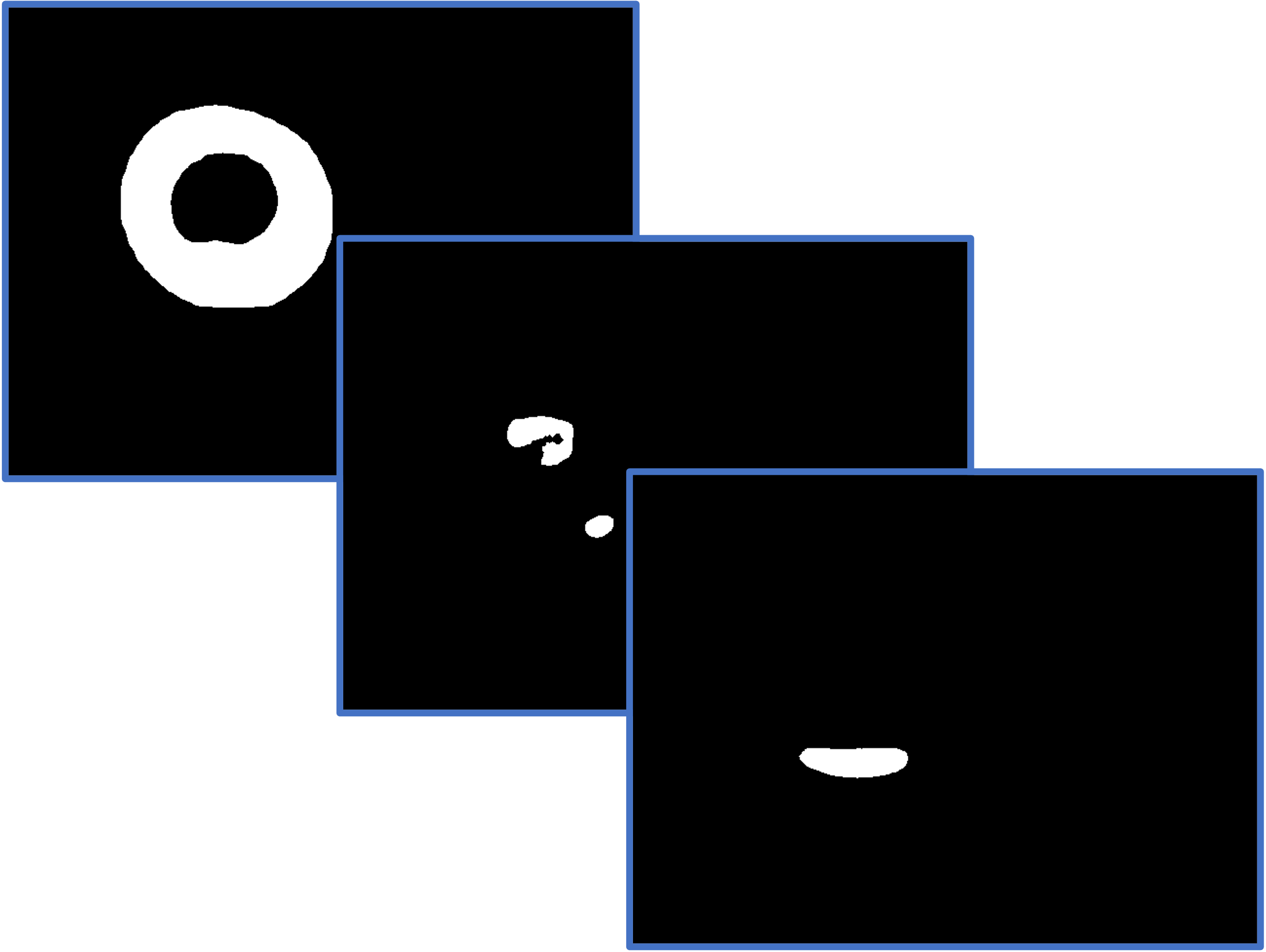}
         \caption{}
     \end{subfigure}\hfill
     \begin{subfigure}{0.22\textwidth}
         \centering
         \includegraphics[width=\textwidth]{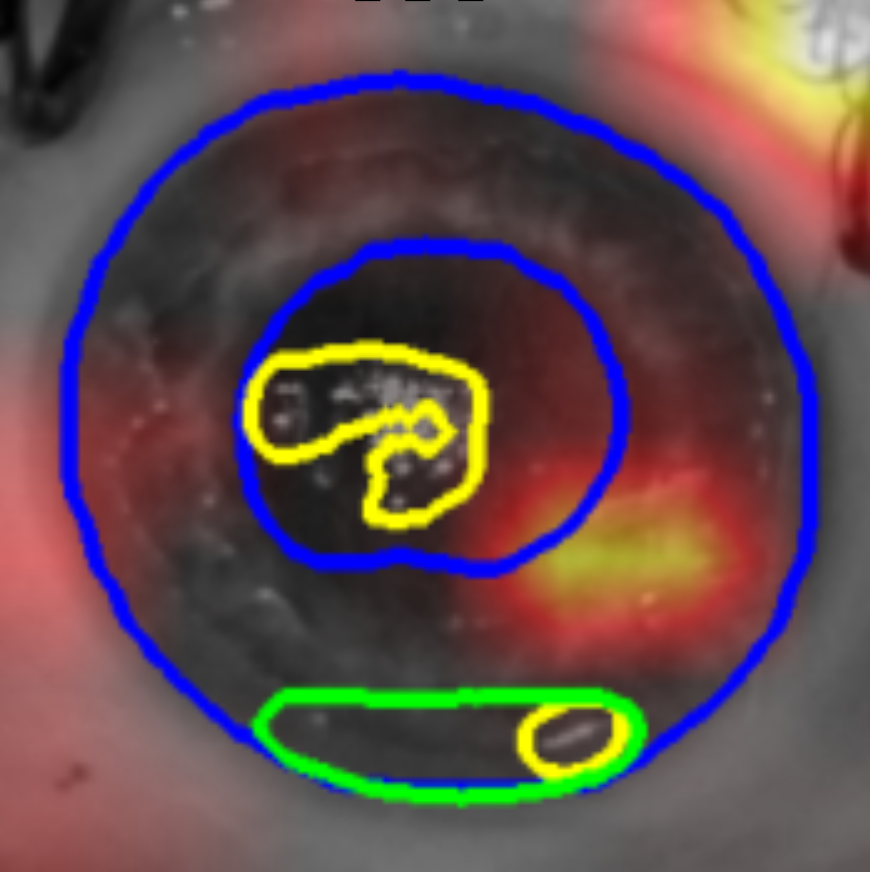}
         \caption{}
     \end{subfigure}
  \caption{Post-mortem iris recognition and visualization: (a) a good-quality post-mortem iris image; (b) top to bottom: deep learning-based detection of iris annulus, specular highlights and decomposition-induced wrinkles; (c) segmentation results presented to a human examiner along with an overlaid heatmap visualizing regions judged as salient by the matching algorithm. Source: \cite{Kuehlkamp_WACV_2022}}
  \label{fig:pmiris}
\end{figure}

Due to decomposition changes to the eye tissues, post-mortem iris images differ significantly from live iris images and rarely meet ISO/IEC 29794-6 quality requirements, as shown in Fig. \ref{fig:pmiris}(a). The challenges are related to appropriate detection of places when cornea dries and generates irregular and large specular highlights, as well regions where iris muscle furrows show up when the eyeball dehydrates. This is where DL-based methods may win over hand crafted approaches, as the latter usually make strong assumptions about anatomy of the iris appearance, not possible to be predicted for eyes undergoing random decomposition processes. Trokielewicz \etal proposed the first known to us iris recognition method designed specifically to cadaver irises \cite{Trokielewicz_WACV2020,Trokielewicz_IMAVIS_2020}. It incorporates SegNet-based segmenter and Siamese networks-based feature extractor, both trained in a domain-specific way solely on post-mortem iris samples. An interesting element of this approach is that segmetation incorporates two models: one trained with ``fine'' ground truth masks, marking all details associated with eye decomposition, and ``coarse'' model, aiming at detecting iris annulus and eyelids, as in classical iris recognition approaches. This allowed to apply a standard ``rubber sheet'' iris images normalization based on ``coarse'' masks, and at the same time exclude decomposition-driven artifacts from encoding, marked by the ``fine'' mask. Kuehlkamp \etal \cite{Kuehlkamp_WACV_2022} in addition to detecting post-mortem deformations, as shown in Fig. \ref{fig:pmiris}(c), they also proposed a human-interpretable visualization of a classification process. The visualization is based on Class Activation Mapping mechanism \cite{Zhou_CVPR_2016} and highlights salient features used by the classifier in its judgment. This novelty in iris recognition algorithms may help human examiners to locate iris regions that should be carefully inspected, or to verify the algorithm's decision.


\section{Human-Machine Pairing to Improve Deep Learning-Based Iris Recognition}
\label{sec:humanpairing}

Iris recognition is usually associated with automatic, solely machine-based and rapid biometric means. It has been changing in the recent decade due to constantly increasing ubiquitousness of iris recognition, especially owing to large governmental applications such as \cite{Aadhaar} or FBI's Next Generation Identification System (NGI) gradually replacing the Integrated Automated Fingerprint Identification System (IAFIS) \cite{FBI_NGI}. This combined with unique identification power of iris whetted the appetite to apply this technique to identification problems normally reserved for fingerprints and face: forensics, lost subjects search or post-mortem identification. To have the legal power, however, the judgment about samples originating or not from the same eye conclusion must be confirmed by a trained human expert. And here is the place where DL-based iris image processing may play a useful role.

Trokielewicz \emph{et al.} compared  iris images in post-mortem iris recognition between humans and machines. They investigated which iris image regions humans and machines mainly attend to compare a pair of images. The machine-based attention maps are generated by Grad-CAM to highlight the regions that contribute the most to the deep learning model's prediction. The human-based attention maps are learned by tracking the gaze as the human is looking around the screen that display iris image pairs and recording the regions where the human spend most time on. Interestingly while humans and machines tend to focus on a limited number of iris areas, however, the region, appearance and density of these areas between humans and machines are different. As salient regions proposed by the deep learning model and identified from human eye gaze do not overlap in general, the computer-added  visual  cues  may  potentially  constitute  a valuable addition to the forensic examiner’s expertise, as it can highlight important discriminatory regions that the human expert might miss in their proceedings. This human-machine pairing is important as human subjects can provide an incorrect decision even despite spending quite sometime  observing  many  iris  regions \cite{NIST_IEG_IETD}. In addition, there has been a body of research showing that humans and machines do not perform similarly well under different conditions \cite{HumanPer,IrisCrypts,Moreira_WACV_2019}. For example, Moreira \emph{et al.} also showed that machines can outperform humans in healthy easy iris image pairs; however, humans outperform machines in disease-affected iris image pairs \cite{Moreira_WACV_2019}. Human-machine pairing will improve deep learning based iris recognition.

\section{Recognition in Less Controlled Environments: Iris/Periocular Analysis}
\label{sec:periocular}

Rooted in the seminal work due to Park \emph{et} al.~\cite{5657254},  efforts have been paid to the development of human recognition methods that - apart the iris - also consider information in the vicinity of the eye to infer the identity. This is a relatively recent topic, termed as \emph{periocular recognition}. The rationale is that the periocular region represents a trade-off between the face and the iris. Periocular biometrics has been claimed to be particularly useful in environments that produce poor quality data (e.g., visual surveillance). Recently, as in the case of iris,  several DL-based solutions have been proposed.

Hernandez-Diaz \emph{et} al.~\cite{8553348}  tested the suitability of off-the-shelf CNN architectures to the periocular recognition task, observing that albeit such networks are optimized to classify generic objects, their features still can be effectively transferred to the periocular domain.   

In the visual surveillance context, Kim \emph{et} al.~\cite{8482116} infer subjects identities based either in loose/tight regions-of-interest, depending of the perceived image quality. Hwang and Lee~\cite{9179802}prevents the loss of mid-level features  and dynamically selects the most important features for classification. Luo \emph{et} al.~\cite{9429938} used self-attention channel and spatial mechanisms into the feature encoding module of a CNN, in order to obtain the most discriminative features of the iris and periocular regions.

 Jung \emph{et} al.~\cite{9159854}'s work is based in the concept of label smoothing regularization (LSR). Having as main goal to reduce the intra-class variability, they described a so-called Generalized LSR (GLSR) by learning a pre-task network prediction that is claimed to improve the permanence of the obtained periocular features. Having similar purposes, Zanlorensi \emph{et} al.~\cite{9191251} described a preprocessing step based in generative networks able to compensate for the typical data variations in visual surveillance environments. Nie \emph{et} al.~\cite{6976788} applied convolutional restricted Boltzmann machines to the periocular recognition problem. Starting from a set of genuine pairs that are used as a constraint, a Mahalanobis distance-metric is learned. 

Obtaining auxiliary (e.g., soft biometrics) has been seen as an interesting direction for compensating the lack of image quality.  Zhao and Kumar~\cite{8353874} incorporate an attention model into a DL-architecture to emphasize the most important regions in the periocular data. The same authors~\cite{7775081} described a semantics-assisted CNN framework to infer comprehensive periocular features. The whole model is composed of different networks, trained upon ID and semantic (e.g., gender, ethnicity) data, that are fused at the score and prediction levels. Similarly, Talreja \emph{et} al.~\cite{9706774} described a multi-branch CNN framework that predicts simultaneously soft biometrics and ID labels, which are finally fused into the final response.
 
With regard to cross-spectral settings, Hernandez-Diaz \emph{et} al. \cite{9304899}  used conditional GANs (CGANs) to convert periocular images between domains, that are further fed to intra-domain off-the-self frameworks. Sharma \emph{et} el.~\cite{7026014} described a shallow neural architecture where each model learns the data features in each spectrum. Then, at a subsequent phase, all models are jointly fine tuned, to learn the cross-spectral variability and correspondence features.

Finally, several works have attempted to faithfully fuse the scores/responses from iris and periocular data. Wang and Kumar~\cite{9194073} used periocular features to adaptively match iris data acquired in less constrained conditions. Their framework incorporates such discriminative information using a multilayer perceptron network. Zhang \emph{et} al.~\cite{Zhang18tifs}  described a DL-model that exploits complementary information from the iris and the periocular regions, that applies \emph{maxout} units to obtain compact representations for each modality and then fuses the discriminative features of the modalities through weighted concatenation. In an opposite direction, Proen\c{c}a and Neves~\cite{8101565} argued that the periocular recognition performance is optimized when the components inside the ocular globe (the iris and the sclera) are simply discarded.

\begin{table*}[h]
    \centering
     \caption{Summary of datasets used in the DL-based iris segmentation and recognition methods of Tables~\ref{tab:Segmentation} and \ref{tab:Recognition} (NIR: \emph{near-infrared}; VW: \emph{visible wavelength}).
     }
     
    \scriptsize
\begin{tabular}{p{3.5cm}p{0.4cm}ccccp{4cm}}
\hline

\textbf{Name} &
\textbf{Data} & 
\textbf{Size} & 
\textbf{\# IDs} & 
\textbf{\# Samples} & 
\textbf{\# Sessions} & 
\textbf{Features} \\
\hline

BATH \cite{Monro07pami} &  NIR &  1280$\times$960  & 1600 & 16000 & 1 &   High quality images \\  \hline

BioSec \cite{Fierrez07PR} &  NIR &  640$\times$480  & 400 & 3200 & 2 &   Office environment \\  \hline

Biosecure \cite{Ortega10pami} &  NIR &  640$\times$480  & 1334 & 2668 & 2 &   Office environment \\  \hline

CASIA-Cross-Sensor \cite{Xiao13btas} &  NIR &  n/a  & 700 & 21000 & 1 &   Multi-sensor, multi-distance (12-30cm, 3-5m) \\  \hline

CASIA-Iris-Distance \cite{Dong09ccpr} &  NIR &  2352$\times$1728  & 284 & 2567 & 1 &   Distant acquisition \\  \hline

CASIA-Iris-Interval \cite{Ma03pami} &  NIR &  320$\times$280  & 395 & 2639 & 2 &   High quality images \\  \hline

CASIA-Iris-Lamp \cite{Wei07icb} &  NIR &  640$\times$480  & 819 & 16212 & 1 &   Non-linear deformation \\  \hline

CASIA-Iris-M1-S1 \cite{Zhang15ccbr} &  NIR &  1920$\times$1080  & 140 & 1400 & 1 &   Mobile device \\  \hline

CASIA-Iris-M1-S2 \cite{Zhang16icb} &  NIR &  1968$\times$1024  & 400 & 6000 & 1 &   Mobile device, multi-distance (20,25,30cm) \\  \hline

CASIA-Iris-M1-S3 \cite{Zhang18tifs} &  NIR &  1920$\times$1920  & 720 & 3600 & 1 &   Mobile device \\  \hline

CASIA-Iris-Thousand \cite{Zhang10icpr} &  NIR &  640$\times$480  & 2000 & 20000 & 1 &  High quality images  \\  \hline

DCME01 \cite{Boyd_Access_2020} &  NIR, VW &  n/a  & 254 & 621 & 1-9 &   - \\  \hline

DCME02 \cite{Kuehlkamp_WACV_2022} &  NIR &  n/a  & 259 & 5770 & 1-53 &   - \\  \hline

IITD \cite{Kumar10pr} &  NIR &  320$\times$240  & 224 & 1120 & 1 &   Varying quality \\  \hline

Iris-Mobile \cite{8987379} & NIR &  n/a  & 750 & 22966 & n/a &   Mobile device, indoor \& outdoor \\  \hline

JluIrisV3.1 \cite{8689110} &  NIR &  640$\times$480  & 120 & 1780 & n/a &   - \\  \hline

JluIrisV4 \cite{8689110} &  NIR &  640$\times$480  & 172 & 114904 & n/a &   - \\  \hline

LivDet-2013-Warsaw \cite{Czajka13mmar} &  NIR &  640$\times$480  & 284 & 1667 & 1 &   High quality images \\  \hline

MICHE-I \cite{Marsico15prl} &  VW &  var.  & 184 & 3732 & 2 &   Three mobile devices \\  \hline

MMU &  NIR &  320$\times$240  & 92 & 460 & 1 &   High quality images \\  \hline

MobBIOfake \cite{Sequeira14ijcb} &  VW &  300$\times$200  & 200 & 1600 & 1 &   With a handheld device \\  \hline

ND-CrossSensor-2013 \cite{Xiao13btas} &  NIR &  640$\times$480  & 1352 & 146550 & 27 &   Multi-sensor \\  \hline

ND-Iris-0405 \cite{ND-Iris-0405} &  NIR &  640$\times$480  & 712 & 64980 & 1 &   Varying quality \\  \hline

ND-TWINS-2009-2010 &  VW &  n/a  & 435 & 24050 & n/a &  Facial pictures frontal, 3/4 and side views.  Indoor \& outdoor \\  \hline


OpenEDS \cite{DBLP:journals/corr/abs-1905-03702} &  NIR &  640$\times$400  & 304 & 356649 & 1 &   From head-mounted VR glasses \\  \hline

Q-FIRE \cite{Johnson10btas} &  NIR &  var.  & 390 & 586560 & 2 &   Iris/face Videos, various distances and quality \\  \hline

UBIRIS.v1 \cite{Proenca05iciap} &  VW &  800$\times$600  & 241 & 1877 & 2 &   Several noise factors \\  \hline

UBIRIS.v2 \cite{Proenca10pami} & VW &  400$\times$300  & 522 & 11102 & 2 &   Distant acquisition, on the move \\  \hline

Warsaw \cite{Boyd_Access_2020} &  NIR, VW &  n/a  & 157 & 4866 & 1-13 &   - \\  \hline

Warsaw-Post-Mortem v1.0 \cite{Trokielewicz_BTAS_2016} &  NIR, VW &  var.  & 34 & 1330 & 2-3 &  Deceased persons, 5-7h to 17 days postmortem \\  \hline

Warsaw-Post-Mortem v2.0 \cite{Trokielewicz_TIFS_2019} &  NIR, VW &  var.  & 73 & 2987 & 1-13 &   Deceased persons \\  \hline


\end{tabular}
    
    \label{tab:segmentation-recognition-databases}
\end{table*}

\section{Open-Source Deep Learning-Based Iris Recognition Tools}
\label{sec:opensources}


Here we summarize the main properties of the datasets employed by the methods of the previous sections for DL-based iris segmentation, recognition and PAD.
%
%
We also describe available open-source software code for these tasks, and other relevant tools.

\subsection{Data Sources}






Table~\ref{tab:segmentation-recognition-databases} gives the 
technical details of the datasets used in the segmentation and recognition methods of Tables~\ref{tab:Segmentation} and \ref{tab:Recognition}. Table~\ref{tab:PAD-databases} does the same for the iris PAD methods of Table~\ref{tab:PAD}.
We show the main properties 
(spectrum, image size, 
identities, images, sessions) and relevant features.
Only the datasets of the methods reported in previous section 
are presented. 
Since we focus on the most recent developments, we consider that such approach provides the most relevant 
datasets for each task.
Of course, the list of available datasets after decades of iris 
research is much longer \cite{Omelina21IVC}.
%


A first observation is the dominance of near infrared (NIR) over the visible (VW) spectrum, 
which should not be surprising, since NIR is regarded as most suitable for iris analysis.
However, research-wise, many segmentation and recognition studies (Tables~\ref{tab:Segmentation}, \ref{tab:Recognition}) use VW images, pushed by the success of challenging databases such as MICHE and UBIRIS. 
On the contrary, the VW modality in iris PAD research is residual (Table~\ref{tab:PAD}), a tendency also observed in pre-DL research \cite{Czajka_ACM_2018,boyd2020iris}.

When it comes to the types of Presentation Attack Instruments (PAIs) employed in iris PAD databases, they can be categorized into:

\begin{itemize}
 
\item PP: paper printout of a real iris image, i.e. from a live person

\item PPD: paper printout of a real iris image with a transparent 3D plastic eye dome on top

\item CLL: textured contact lenses worn by a live person

\item CLP: textured contact lenses on printout (either a printout of a CLL image, or a printout of a real iris image with a textured contact lens placed on top)

\item RA: replay attack, i.e. a real iris image shown on a display

\item AE: artificial eyeball (plastic eyes of two different types: Van Dyke Eyes, with higher iris quality details, and Scary eyes, plastic fake eyes with a simple pattern on the iris region)

\item AEC: artificial eyeball with a textured contact lens on top

\item SY: synthetic iris, i.e. an image created via generative methods

\item PM: postmortem iris, i.e. an image acquired from cadaver eyes

\end{itemize}

These PAIs mostly entail presenting the mentioned instrument to the iris sensor, which then captures an image of the 
artifact.
An exception is ``SY'', which directly produces a synthetic digital 
image, 
although such image could be used as base to, 
for example, PP, PPD, RA, or AE attacks.
In Table~\ref{tab:PAD-databases}, it can be seen that CLL (textured lenses live) and PP (paper printouts) largely dominates as the most popular PAIs on the existing databases, and consequently, on the related research (Table~\ref{tab:PAD}).
CLP (textured lenses on printout) also appears in many studies, driven by the wide use of the LivDet-2017-IIITD-WVU set, which includes such PAI.
CASIA-Iris-Fake, which contains AE (artificial eyes) and SY (synthetic irises) also appears in a few 
studies.
Other attacks that one may expect on the digital era, such as RA (replay), however, are residual in datasets and recent studies.

\begin{table*}[h]
    \centering
     \caption{Summary of datasets used in the DL-based iris Presentation Attack Detection methods of Table~\ref{tab:PAD} (NIR: \emph{near-infrared}; VW: \emph{visible wavelength}).
     The type of PAIs (second column) are PP: paper printout, PPD: paper printout with plastic dome, CLL: textured contact lenses (live), CLP: textured contact lenses (printout), RA: replay attack (display), AE: artificial eyeball, AEC: artificial eyeball with textured contact lens, SY: synthetic iris, PM: postmortem iris.
     TTP (next to last column) indicates the existence of a training/test split.
     The features (last column) are MS: multi-sensor, ME: multi-environment (e.g. indoor/outdoor, light variability, mobile environment, etc.), UPAI: unseen PAIs in the test set.
     }
     
    \scriptsize
\begin{tabular}{p{2cm}p{1.4cm}p{0.4cm}ccccccp{0.4cm}p{3cm}}
\hline

\multirow{2}{*}{\textbf{Name}} &
\multirow{2}{*}{\textbf{PAIs}} &
\multirow{2}{*}{\textbf{Data}} &
\multirow{2}{*}{\textbf{Size}} &
\multicolumn{2}{c}{\textbf{\# IDs}} & 
\multicolumn{3}{c}{\textbf{\# Samples}} & 
\multirow{2}{*}{\textbf{TTP}} & 
\multirow{2}{*}{\textbf{Features}} \\
\cline{5-9}

 &  &  &  & \textbf{live} & \textbf{fake} & \textbf{live} & \textbf{fake} & \textbf{total} &  &  \\ 
\hline

CASIA-Iris-Fake \cite{Sun14pami} & PP, CLL, AE, SY & NIR &  640$\times$480  & 1000 & 815 & 6000 & 4120 & 10240 &  &  \\ 
\hline


IF-VE \cite{Zhang20ICPR} & CLL  & NIR &  n/a  & 200 & 200 & 25000 & 25000 & 50000 & \cmark  &   MS, ME \\ 
\hline

IPITRT \cite{Peng20aeeca} & PP & NIR &  var.  & 58 & n/a & 1800 & 551 & 2351 &  &  ME \\ 
\hline

IIITD-CLI \cite{Kohli13icb} & CLL & NIR &  640$\times$480  & 202 & n/a & n/a & n/a & 6570 & \cmark &  MS \\ 
\hline

IIITD-IS$^3$ \cite{Gupta14icpr} & PP, CLP & NIR &  640$\times$480  & 202 & n/a & 0 & 4848 & 4848 &  &   MS \\ 
\hline

\multicolumn{9}{l}{LivDet-2017 \cite{Yambay17livdetIJCB}} \\

-Clarkson  & PP, CLL & NIR &  640$\times$480  & 50 & n/a & 3954 & 4141 & 8095 &  \cmark & UPAI (additional patterned lenses) \\ 
\cline{2-10}

-IIITD-WVU$^1$ & PP, CLL, CLP & NIR &  640$\times$480  & n/a & n/a & 2952 & 4507 & 7459 & \cmark & MS, ME, UPAI (additional patterned lenses) \\ 
\cline{2-10}

-ND-CLD$^2$  & CLL & NIR &  640$\times$480  & n/a  & n/a  & 2400 & 2400 & 4800 & \cmark &  UPAI (additional patterned lenses) \\ 
\cline{2-10}

-Warsaw & PP & NIR &  640$\times$480  & 457 & 446 & 5168 & 6845 & 12013 & \cmark &  MS \\ 
\hline

LivDet-2020 \cite{Das20livdetIJCB}  & PP, PPD, CLL, CLP, RA, AE, AEC, PM & NIR &  640$\times$480  & n/a & n/a & 5331 & 7101 & 12432 &  &  MS \\ 
\hline

Iris-CL1 \cite{Tapia22TIFS} & PP & NIR &  var. & n/a & n/a & n/a & 1800 & n/a &  &   MS \\  \hline

JHU-APL \cite{Chen21wacv} & CLL, AE & NIR &  n/a  & n/a & n/a & 7191 & 7214 & 14405 &  &   ME \\ 
\hline

MSU-IrisPA-01 \cite{Yadav19CVPRW} & PP, CLL, RA, AE & NIR &  640$\times$480  & n/a & n/a & 1343 & 2523 & &   &   \\ 
\hline

MUIPA \cite{Yadav18wacv} & PP, CLL  & NIR &  640$\times$480  & 70 & 70 & n/a & n/a & 10296 &   &  ME \\ 
\hline

ND-CLD-13 \cite{Doyle13btas} & CLL & NIR &  640$\times$480  & 330 & n/a & 3400 & 1700 & 5100 & \cmark &   MS \\ 
\hline

ND-CLD-15$^2$ \cite{Doyle15access} & CLL & NIR &  640$\times$480  & n/a & n/a & 4800 & 2500 & 7300 & \cmark  &  MS \\ 
\hline

NDIris3D \cite{Fang21tifs} & CLL & NIR &  640$\times$480  & 176 & 176 & 3458 & 3392 & 6850 &  &   MS \\ 
\hline

ND-PSID$^4$ \cite{Czajka19wacv} & CLL &  NIR &  640$\times$480  & 238 & 238 & 3132 & 2664 & 5796 &  &   \\ 
\hline

UnMIPA \cite{Yadav19acvprw} & CLL  & NIR &  640$\times$480  & 162 & 162 & 9319 & 9387 & 18706 &   &  MS, ME \\ 
\hline

Warsaw-Post-Mortem v3.0 \cite{Trokielewicz_IMAVIS_2020} &  PM   & NIR, VW &  var.  & 0 & 79 & 0 & 1879 & 1879 &   &  MS \\ 
\hline



\multicolumn{9}{l}{$^1$ Contains IIITD-CLI and IIITD-IS} \\

\multicolumn{9}{l}{$^2$ Iris-LivDet-2017-ND-CLD is a subset of ND-CLD-15} \\

\multicolumn{9}{l}{$^3$ IIITD-IS images are printouts of IIITD-CLI captured with a iris scanner and a flatbed scanner} \\

\multicolumn{9}{l}{$^4$ ND-PSID is a subset of ND-CLD-15} \\

\end{tabular}
    
    \label{tab:PAD-databases}
\end{table*}

\subsection{Software Tools}

%
%
%

The availability of DL-based tools for iris biometrics has been scarce for years, specially for PAD \cite{Fang21tifs}. In the following, we provide a short description of peer-reviewed references with associated source code 
(link included in the paper, or easily found on the websites of the authors or dedicated sites such as www.paperswithcode.com). We describe (in this order) tools for segmentation, recognition and PAD. For each type, the references are then presented in cronological order.

\subsubsection{Segmentation} \hfill

Lozej \textit{et al}. \cite{Lozej18iwobi} released their end-to-end DL model 
based on the U-Net architecture \cite{U-Net}.
The model was trained and evaluated with a small set of 200 
annotated iris images from CASIA database.
%
The authors also explored the impact of the model depth and the use of batch normalization layers. 

Kerrigan \textit{et al}. \cite{8987299} released the code and models of Iris-recognition-OTS-DNN, a set of four architectures based on off-the-shelf CNNs trained for iris segmentation (two VGG-16 with dilated convolutions, one ResNet with dilated kernels, and one SegNet encoder/decoder).
%
%
Training databases included CASIA-Irisv4-Interval, ND-Iris-0405, Warsaw-Post-Mortem v2.0 and ND-TWINS-2009-2010, whereas testing data came from ND-Iris-0405 (disjoint subject), BioSec and UBIRIS.v2.
Results showed that the DL solutions evaluated outperform traditional segmentation techniques, e.g. Hough transform or integro-differential operators.
It was also seen that each test dataset had a method that performs best, with UBIRIS obtaining the worst performance. This should not come as a surprise, since it contains VW images with high variability taking distantly with a digital camera, whereas the other two are from close-up NIR iris sensors in controlled environments.

Wang \textit{et al}. \cite{IrisParseNet} released the code and models of their high-efficiency segmentation approach, 
IrisParseNet. 
%
A multi-task attention network was first applied to simultaneously predict the iris mask, pupil mask and iris outer boundary. 
Then, from the predicted masks and outer boundary, a parameterization of the iris boundaries was calculated. 
The solution is complete, in the sense that the mask (including light reflections and occlusions) and the parameterized inner and outer iris boundaries are jointly achieved.
%

More recently, authors from the same group presented IrisSegBenchmark \cite{Wang20JCRD}, an open iris segmentation evaluation benchmark where they implemented 
six different CNN architectures, including
Fully Convolutional Networks (FCN) \cite{FCN}, 
Deeplab V1,V2,V3 \cite{DBLP:journals/corr/ChenPK0Y16},
ParseNet \cite{Liu16iclr},
PSPNet \cite{Zhao17cvpr},
SegNet \cite{Badrinarayanan_TPAMI_2017}, and
U-Net \cite{U-Net}.
The methods were evaluated on CASIA-Irisv4-Distance, MICHE-I and UBIRIS.v2.
As in \cite{8987299}, results showed that the best method depends on the database, being: ParseNet for CASIA (NIR data), DeeplabV3 for MICHE (VW images from mobile devices), and U-Net for UBIRIS (VW images from a digital camera).
In this case, however, the three tests databases behaved approximately equal, since they all contain difficult distant data. CASIA showed a slightly better accuracy, suggesting that NIR data may be easier to segment.
Traditional, non-DL methods were also evaluated, concluding that DL-based segmentation achieves superior accuracy.

Banerjee \textit{et al}. \cite{Banerjee22SNCS} released the code of their V-Net architecture,  
designed to overcome some drawbacks of U-Net, such as instability to tackle iris segmentation or tendency to overfit.
A pre-processing stage on the YCrCb and HSV spaces was also added to detect salient regions and aid detection of iris boundaries.
The method was evaluated on the difficult UBIRIS.v2 VW dataset.

\subsubsection{Recognition} \hfill

The code of the DL method 
ThirdEye was released by Ahmad and Fuller \cite{Ahmad19btas}, based on a ResNet-50 trained with triplet loss. 
 Authors directly used segmented images without normalization to a rectangular 2D representation, arguing that such step may be counterproductive in unconstrained images. 
The model was evaluated on the ND-0405, IITD and UBIRIS.v2 datasets.
%

The models of Boyd \textit{et al}. \cite{DeepFeatExtr} for recognition have been also released, based on
%
a ResNet-50 with different weight initialization techniques, comprising: from scratch (random), off-the-shelf ImageNet (general-purpose vision weights), off-the shelf VGGFace2 (face recognition weights), fine-tuned ImageNet weights, and fine-tuned VGGFace2 weights.
Both ImageNet and VGGFace2 are very large datasets with millions of images, and face images contain the iris region. Thus, using these datasets as initialization may be beneficial for iris recognition, where available training data is in the order of hundreds of thousand images only.
This strategy has been followed e.g. in ocular soft-biometrics as well \cite{Alonso21ietb}.
The observed optimal strategy is indeed to fine-tune an off-the-shelf set of weights to the iris recognition domain, be general-purpose or face recognition weights. 

\subsubsection{Segmentation and Recognition Packages} \hfill

A complete package comprising segmentation and feature encoding was provided by Tann \textit{et al}.\cite{tann2019resource}.
The segmentator is based on a Fully Convolutional Network (FCN), but encoding is based on hand-crafted Gabor filters \cite{IrisCode}.
Evaluation was done on CASIA-Irisv4-Interval and IITD.
%

%
In forensic investigation for diseased eyes and post-mortem samples, Czajka \cite{Czajka21githubIris} also released a complete package combining segmentation and feature encoding. 
The models are based on previous efforts of the author and co-workers, comprising a SegNet \cite{Trokielewicz_IMAVIS_2020} and a CCNet \cite{CC-Net} DL segmentators, but the feature encoder is based on hand-crafted BSIF filters.
%

%
Another complete segmentation and recognition package was released by Kuehlkamp \textit{et al}. \cite{Kuehlkamp_WACV_2022}. The segmentator is based on a fine-tuned Mask-RCNN architecture, with the cropped iris region fed directly into a ResNet50 pre-trained for face recognition on the very large VGGFace2 dataset, and fine-tuned for iris recognition using triplet loss.
The paper is oriented towards postmortem iris analysis, so the methods use a mixture of live and postmortem images for training and evaluation.

Parzianello and Czajka \cite{Parzianello22WACVW} also released the models and annotated data for their textured contact lens aware iris recognition method. The foundation is that such lenses may be used normally for cosmetic purposes, without intention to fool the biometric system. Therefore, they proposed to detect and match portions of live iris tissue still visible in order to enable recognition even when a person wears textured contact lenses.
To do so, they applied a Mask R-CNN as a segmentation backbone, trained to detect authentically-looking parts of the iris using manually segmented samples from NDIris3D dataset.
Non-iris information is then removed from the training images by blurring it or replacing it with random noise to guide the subsequent recognition network (based on ResNet-18) to salient, non-occluded regions that should be used for matching.

\subsubsection{Iris PAD} \hfill

In the iris PAD arena, Gragnaniello \textit{et al}. \cite{Gragnaniello16sitis} proposed a CNN that incorporates domain-specific knowledge. 
Based on the assumption that PAD relies on residual artifacts left mostly in high-frequencies, 
a regularization term was added to the loss function which forces the first layer to behave as a high-pass filter.
The method, which is available in the website of the first author, could be applied to PAD in multiple modalities, including iris and face.

The code and model of the method of Sharma and Ross \cite{Sharma20ijcb} (D-NetPAD) is also available. 
It is based on DenseNet121 and trained for a variety of PAIs (printouts, artificial eye, cosmetic contacts, kindle replay, and transparent dome on print), with an script to retrain the method also available.

\subsection{Other Tools: Iris Image Quality Assessment}


%
Several image properties considered to potentially influence the accuracy of iris biometrics have been defined in support of the standard ISO/IEC 29794-6 \cite{ISO/IEC29794-6}.
They include: grayscale spread (dynamic range), iris size (pixels across the iris radius when the boundaries are modeled by a circle), dilation (ratio of the pupil to iris radius), usable iris area (percentage of non-occluded iris, either by eyelashes, eyelids or reflections), contrast of pupil and sclera boundaries, shape (irregularity) of pupil and sclera boundaries, margin (distance between the iris boundary and the closest image edge), sharpness (absence of defocus blur), motion blur, signal to noise ratio, gaze (deviation of the optical axis of the eye from the optical axis of the camera), and interlace of the acquisition device.

Low quality iris images, which can potentially appear in uncontrolled or non-cooperative environments, are known to reduce the performance of iris location, segmentation and recognition. Thus, an accurate quality assessment can be a valuable tool in support of the overall pipeline, either by dropping low quality images, or invoking specialized processing \cite{6095497}. 
One possibility might be to quantify the properties mentioned above, and placing thresholds on each. 
A more elaborated alternative is to combine them according to some rule and produce an overall quality score. However, it is difficult to provide metrics that cover all types of quality distortions \cite{[Tabassi11]} and doing so for some indeed entails to segment the iris.

Broadly, a biometric sample is of good quality if it is suitable for recognition, so quality should correlate with recognition performance \cite{[Grother07]}.
As such, quality assessment can be viewed as a regression problem.
Wang \textit{et al}. \cite{Wang20ijcb} 
considered that a non-ideal eye image will pivot in the feature space around the embedding of an ideal image. 
They defined quality as the distance to the embedding of such ``ideal'' image which, is regarded as a registration sample collected under a highly controlled environment.
%
They used a model to learn the mapping between images and Distance in Feature Space (DFS) directly from a given dataset.
Quality is computed via attention-based pooling that combines a heatmap that comes from a coarse segmentation based on U-Net and the feature map of an extraction network based on MobileNetv2 pre-trained on CASIA-Iris-V4 and NDIRIS-0405.
%

\section{Emerging Research Directions}
\label{sec:open}

 In this section, we discuss the most relevant open challenges and hypothesize about emerging research directions that could become \emph{hot-topics} in biometrics literature in a close future.

\subsection{Resource-aware designs of iris recognition networks} 
Application-wise, iris recognition can be performed on a wide range of hardware, ranging from high-end computers to low-end embedded devices, or from large computer clusters to personal devices such as mobile phones. Performing recognition on resource-limited hardware could pose new challenges for deep learning based iris networks, which usually contain hundreds of layers and millions of parameters. Therefore designing these deep learning networks necessarily need to be aware of the hardware platforms on which they will be run.

\vspace{3px}
\noindent\textit{Lightweight models:}
Lightweight CNNs employ advanced techniques to efficiently trade-off between resource and accuracy, minimising their model size and computations in term of the number of floating point operations (FLOPs), while retaining high accuracies. Specialized lightweight CNN architectures include MobileNets \cite{MobileNetV3} and U-Net \cite{U-Net}. 
There are a few lightweight deep learning based models for both segmentation and feature extraction. Fang \emph{et al.} \cite{Fang_IJCB_2020} adapted the lightweight CC-Net \cite{CC-Net} for iris segmentation. CC-Net has a U-Net structure \cite{U-Net}, able to  retain up to 95\% accuracy using only 0.1\% of the trainable parameters. Boutros \emph{et al.} \cite{Boutros2020} benchmarked MobileNet-V3 against deeper networks for iris recognition and showed that the MobileNet based model can achieve similar EER with 85\% less number of parameters and 80\% less inference time.

\vspace{3px}
\noindent\textit{Model compression:}
Studies have found that most of the large deep learning models tend to be overparameterized, leading to lots of
redundant parameters and operations in the network. This becomes more severe considering iris texture images are different from generic object-based images. This has
motivated a hot trend looking to remove these redundancies
from the models, including pruning, quantization and low-rank factorization \cite{PruningQuantiSurvey}. 
In our iris recognition literature, there a few lightweight deep learning based models for both segmentation and feature extraction. Tann \emph{et al.} \cite{tann2019resource} quantized 64-bit floating points numbers of weights and activations of the full FCN-based iris segmentation model using an 8-bit dynamic fixed-point (DFP) format, which provide a 8$\times$ memory saving as well as speed enhancement due to reduced complexity of lower precision operations.

\vspace{3px}
\noindent\textit{Neural Architecture Search:}
Neural Architecture Search (NAS) automates the process of architecture design of neural networks by iteratively sampling a population of child networks, evaluating the child models’ performance metrics as rewards and learning to generate high-performance architecture candidates \cite{elsken2019neural}. 
In our iris recognition literature, Nguyen \emph{et al.} \cite{ConstrainedIrisNet} showed that computation and memory can be incorporated into the NAS formulation to enable resource-constrained design of deep iris networks.

\subsection{Human-interpretable methods and XAI} 
With hundreds of layers and millions of parameters, deep learning networks are usually opaque or ``blackbox'' where humans struggle to understand why a deep network predict what it predicts. This necessitates approaches to make deep learning methods more interpretable and understandable to humans. Interestingly, the need for human-interpretable methods has been raised even from the handcrafted era. For example, Shen \emph{et al.} published a series of work \cite{IrisCrypts,IrisCrypts1} on using iris crypts for iris matching. Iris crypts are clearly visible to humans in a similar way as finger minutiae. Another example is the macro-features \cite{IrisKeypoints} which use SIFT to detect keypoints and perform iris matching based on these keypoints~\cite{NIST_VIfeatures}. Another notable work is by Proença \emph{et al.} \cite{IRINA} where they proposed a deformation field to represent the correspondence between two iris images.
  
From a deep learning perspective, researchers have also attempted to visualize the matching. Kuehlkamp \emph{et al.} \cite{Kuehlkamp_WACV_2022} argued that existing iris recognition methods offer limited and non-standard methods of visualization to let human examiners interpret the model output. They applied Class Activation Maps (CAM) \cite{Zhou_CVPR_2016} to visualize the level of contribution of each iris region to the overall matching score. Similarly, Nguyen \emph{et al.} \cite{ComplexIrisNet} also decomposed the final matching score into pixel-level to visualize the level of contribution of each pixel to the overall matching score. 


\subsection{Deep learning-based synthetic iris generation} 


Data synthesis
provides an alternative to time- and resource-consuming database collection. 
%
One could create as many images as desired, with new textures that even do not match any existing identity, which would avoid privacy problems too.
On the other hand, fake irises that are indistinguishable from real ones can be used for identity concealment attacks (if the image does not match any identity) or impersonation attacks (if the image resembles an existing identity) \cite{Czajka_ACM_2018}. Indeed, synthetic irises are present in databases employed for iris PAD, such as CASIA-Iris-Fake (Table~\ref{tab:PAD-databases}). 

Regardless of the purpose or ability to detect if an image is synthetic, Generative Adversarial Networks (GANs) \cite{10.5555/2969033.2969125} have shown impressive photo-realistic generating capabilities in many domains. GANs learn to model image distributions by an adversarial process, where a discriminator 
assesses the realism of images synthesized by a generator.  
At the end, the generator have learned the 
distribution of the training data, being able to synthesize new images with the same characteristics. 

For iris generation, some methods by Yadav \textit{et al}. \cite{Yadav21WACV,Yadav20WACV,Yadav19CVPRW} were mentioned in iris PAD contexts (Section~\ref{sec:PAD-adversarial}).
RaSGAN \cite{Yadav20WACV,Yadav19CVPRW} followed the traditional approach of driving the generation/discrimination training by randomly sampling so-called latent vectors from a probabilistic distribution. As training progresses, the generator learns to associate features of the latent vectors with semantically meaningful attributes that naturally vary in the images. 
However, this does not impose any restriction in the relationship between features in latent space and factors of variation in the image domain, making difficult to decode what the latent vectors represent.
As a result, the image characteristics (eye color, eyelids shape, eyelashes, gender, age...) are generated randomly. 
%
%
%
Kohli \textit{et al}. \cite{Kohli17ijcb} presented iDCGAN for iris PAD, which also followed the latent vector sampling concept. 
To counteract such issue, researchers have tried to incorporate constrains or mechanisms that guide the generation process to a desired characteristic. For example, CIT-GAN \cite{Yadav21WACV} employed a Styling Network that learns style characteristics of each given domain, while taking as input a domain label that drives the network to embed a desired style into the generated data.

In a similar direction, Kaur and Manduchi \cite{Kaur21wacv,Kaur20wacv} proposed to synthesize eye images with a desired style (skin color, texture, iris color, identity) using an encoder-decoder ResNet. The method is aimed at manipulating gaze, so the generator receives a segmentation mask with the desired gaze, and an image with the style that will see its gaze modified. 
To achieve cross-spectral recognition, Hernandez-Diaz \textit{et al}. \cite{9304899} used CGANs to convert ocular images between VW and NIR spectra while keeping identity, so comparisons are done in the same spectrum. This allows the use of existing feature methods, which are typically optimized to operate in a single spectrum.

Despite great advances in DL-based synthetic image generation, one open problem is the possible identity \textit{leakage} 
from the training set when creating data of non-existing identities, resulting in privacy issues. This has just been revealed recently in face generation \cite{Tinsley_WACV_2021}.
Another issue in the opposite direction is the difficulty in preserving identity in the generation process when the target is precisely creating images of an existing identity with different properties.
This is an issue being addressed in face generation methods [reference under review], but is lacking in iris synthesis research.

\subsection{Deep learning-based iris super-resolution} 
One of the main constraints for existing iris recognition systems is the short distance of image acquisition, which usually requires a subject to stay still less than 60 cm from iris cameras. This is due to the requirement of high-resolution iris region, \emph{e.g.} 120 pixels across the iris diameter due to the European standard and NIST standard, despite the small physical size of an eye, \emph{i.e.} $15\times15$ mm. The lack of resolution of imaging systems has critically adverse impacts on the recognition and performance of biometric systems, especially in less constrained conditions and long range surveillance applications \cite{SRbiometrics}. 

Super-resolution, as one of the core innovations in computer vision, has been an attractive but challenging solution to address the low resolution problem in both general imaging systems and biometric systems. Deep learning based super-resolution approaches have been across multiple works in iris recognition. Ribeiro \emph{et al.} \cite{DL_SRiris,RealisticSRiris} experimented two deep learning single-image super-resolution approaches: Stacked Auto-Encoders (SAE) and Convolutional Neural Networks (CNN). Both approaches learn one encoder to map the high resolution iris images to the low resolution domain, and one decoder to learn to reconstruct the original high resolution images from the low resolution ones. Zhang \emph{et al.} \cite{Mobile_SRiris} learned a single CNN to learn non-linear mapping function between LR images to HR images for mobile iris recognition. Wang \emph{et al.} \cite{GAN_SRiris} extended the single CNN to two CNNs: one generator CNN and one discriminator CNN as in the GAN architecture. The generator functions similar to the single LR - HR mapping CNN. Adding the discriminator CNN allows them to control the generator to generate HR images not just visually higher resolution but also preserve the identity of the iris. Mostofa \emph{et al.} \cite{xGANiris} incorporated a GAN-based photo-realistic super-resolution approach \cite{SRGAN} to improve the resolution of LR iris images from the NIR domain before cross-matching the HR outputs with the HR images from the RGB domain. While these approaches showed improved performance, dealing with noisy data in such cases as iris at a distance and on the move could require the quality of an input iris image to be included in the super-resolution process \cite{QualitySR}. In addition, Nguyen \emph{et al.} argued that a fundamental difference exists between conventional super-resolution motivations and those required for biometrics, hence proposing to perform super-resolution at the feature level targeting explicitly the representation used by recognition \cite{FeatureSR}.

\subsection{Privacy in deep learning-based iris recognition} 

Privacy is becoming a key issue in computer vision and machine learning domains. In particular, it is accepted that the accuracy attained by deep learning models depends on the availability of large amounts of visual data, which stresses the need for privacy-preserving recognition solutions. 

In short, the goal in privacy preserving deep-learning is to appropriately train models while preserving the privacy of the training datasets. While the utility of this kind of solutions is obvious, there are certain concerns about the training data that supported the model creation, as the collection of images from a large number of  individuals comes with significant privacy risks. In particular, it should be considered that the subjects from whom the data were collected can neither delete nor control what actually will be learned from their data.

As most of the existing biometric technologies, DL-based iris recognition pose challenges to privacy, which are even more concerning, considering the \emph{data-driven} feature of such kind of systems. Particular attention should be paid to avoid function creep, guaranteeing that the system yielding from a set of data is not used for a different purpose than the originally communicated to the individual at the time of providing their information. Covert collection is another major concern, which is also particular important for the iris trait, according to the possibility of being imaged from large distances and in surreptitious way.

Particular attention has been paid to the development of fair recognition systems, in the sense that this kind of systems should attain similar effectiveness in different subgroups of the population, regarding different features such as \emph{gender}, \emph{age}, \emph{race} or \emph{ethnicity}. For data-driven systems, this might be a relevant challenge, considering that most of the existing datasets that support the learned systems have evident biases with regared tio the subjects' characteristics above. 

Lastly, in a more general machine learning perspective, potential attacks to the learned models have been concerning the research community and have been the scope of various recent works, attempting to provide defense mechanisms against: i) model inversion attacks, that aim to reconstruct the training data from the model parameters (e.g.,~\cite{9672188} and~\cite{9252914}); ii) membership inference, that attempt to infer whether one individual was part of a training set (e.g.,~\cite{9773984} and~\cite{8844607}); and iii) training data extraction attacks, that aim to recover individual training samples by querying the models (e.g,~\cite{8854425} and~\cite{9099993}).

\subsection{Deep learning-based iris segmentation} 

Being one of the earliest phases of the recognition process, segmentation is known as one of the most challenging, as it is at the front line for facing the dynamics of the data acquisition environments. This is particularly true, in case of less constrained data acquisition protocols, where the resulting data have highly varying features and the particular conditions of each environment strongly determine the most likely data covariates. 

In the segmentation context, the main challenge remains as the development of methods robust to \emph{cross-domain} settings, i.e., able to segment the iris region for a broad range of image features, e.g., in terms of: 1) illumination, 2) scale, 3) gaze, 4) occlusions, 5) rotation and 6) pose, corresponding to the acquisition in very different environments.  Over the past decades, many research groups have been devoting their attentions in improving the robustness of iris segmentation, which is known to be a primary factor for the final effectiveness of the recognition process. In this timeline, the proposed segmentation methods can be roughly grouped into three categories: 1) boundary-based methods (using the integro-differential operator or Hough transform); 2) based in handcrafted features (particularly suited for non-cooperative recognition, e.g.,~\cite{TAN2010223} and~\cite{6199979}) ; and 3) DL-based solutions.

For the latter family of methods, the emerging trends are closely related to the \emph{general} challenges of DL-based segmentation frameworks, namely to obtain interpretable models that allow us to perceive what exactly are these systems learning, or the minimal neural architecture that guarantees a predefined level of accuracy. Also, the development of weakly supervised or even unsupervised frameworks is another \emph{grand-challenge}, as it is accepted that such systems will likely adapt better to previously unseen data acquisition conditions. Finally, the computational cost of segmentation (both in terms of space and time) is another concern, with special impact in the deployment of this kind of frameworks in mobile settings, and in the IoT setting~\cite{9821888}.

\subsection{Deep learning-based iris recognition in visible wavelengths} 

Being a topic of study for over a decade (e.g.~\cite{LIU201966} and~\cite{Proenca2013}), iris recognition in visible wavelengths remains essentially as an interesting possibility for delivering biometric recognition from large distances (in conditions that are typically associated to visual surveillance settings) and in handheld \emph{commercial} devices, such as smartphones.

The emerging trends in this scope regard the development of alternate ways to analyze the multi-spectral information available in visible light data (typically RGB), i.e., by developing deep learning architectures optimized for fusion, either at the data, feature, score or decision levels~\cite{BIGDELI2021107563}.

In the visual surveillance setting, the main challenge regards the development of optimized data acquisition settings, profiting from the advances in remote sensing technologies, that should be able to augment the quality (e.g., resolution and sharpness) of the obtained irises. In this scope, the research on active data acquisition technologies (based in PTZ devices, or similar) might also be an interesting emerging possibility~\cite{9421103}.

\section{Conclusions}
\label{sec:Conclusion}


Motivated by the tremendous success of DL-based solutions for many different solutions to everyday problems, machine learning is entering one of its golden era, attracting growing interests from the research, commercial and governmental communities.  In short, deep learning uses multiple layers to represent the abstractions of data to build computational models that - even in a bit surprising way - typically surpass the previous generation of handcrafted-based automata. However, being extremely data-driven, the effectiveness of DL-based solutions is typically constrained by the existence of massive amounts of data, annotated in a consistent way.  

As in the generality of the computer-vision topics, a myriad of DL-based techniques has been proposed over the last years to  perform biometric recognition, and - in particular - iris recognition. Nowadays, the existing methods cover the whole phases of the typical processing chain, from the preprocessing, segmentation, feature extraction up to the matching and recognition steps. 

Accordingly, this article provides the first comprehensive review of the historical and state-of-the-art approaches in DL-based techniques for iris recognition, followed by an in-depth analysis on pivoting and groundbreaking advances in each phase of the processing chain. We summarize and critically compare the most relevant methods for iris acquisition, segmentation, quality assessment, feature encoding, matching and recognition problems, also presenting the most relevant open-problems for each phase.

Finally, we review the typical issues faced in DL-based methods in this domain of expertize, such as unsupervised learning, black-box models, and online learning and to illustrate how these challenges can be important to open prolific future research paths and solutions.

%

\begin{acks}
We would like to thank Adam Czajka from the University of Notre Dame, USA for the contribution in the early version of this survey paper in Sections 1, 5 and 6.
The work due to Hugo Proen\c{c}a was funded by FCT/MEC through national funds and co-funded by FEDER - PT2020 partnership agreement under the projects UIDB/50008/2020, POCI-01-0247-FEDER- 033395.
Author Alonso-Fernandez thanks the Swedish Innovation Agency VINNOVA (project MIDAS and DIFFUSE) and the Swedish Research Council (project 2021-05110) for funding his research.

\end{acks}

%
\bibliographystyle{ACM-Reference-Format}
\bibliography{DLIR}

%
\appendix

\end{document}